\documentclass[10pt,twocolumn]{article}
\usepackage{cvpr}
\usepackage{times}
\usepackage{epsfig}
\usepackage{amsmath,amssymb,amsfonts,amsthm,bm}
\usepackage{bbm}
\usepackage{caption}
\usepackage{mwe, subfig, array, calc}

\renewcommand{\thesubfigure}{\alph{subfigure}}
\newcommand{\mycaption}[1]
{\refstepcounter{subfigure}\textbf{(\thesubfigure) }{\ignorespaces #1}}
\newlength{\tempheight}
\newlength{\tempwidth}
\newcommand{\rowname}[1]
{\rotatebox{90}{\makebox[\tempheight][c]{#1}}}
\newcommand{\columnname}[1]
{\makebox[\tempwidth][c]{#1}}%
\usepackage{graphicx}
\usepackage{textcomp}
\usepackage{array}
\usepackage{multirow}
\usepackage{booktabs}
\usepackage{tabularx}
\usepackage{rotating}
\usepackage{stfloats}
\usepackage[pdftex,pagebackref=true,bookmarks=false]{hyperref}
\usepackage{ml-acro}
\usepackage{balance}
\usepackage[table,xcdraw]{xcolor}
\usepackage{booktabs}

\def\cvprPaperID{2046} 

\cvprfinalcopy

\newcommand{\mtilde}{\raise.17ex\hbox{$\scriptstyle\sim$}}
\usepackage{pifont}
\newcommand{\cmark}{\ding{51}}%
\title{Learning to Fuse Things and Stuff}
\author{Jie Li*, Allan Raventos*, Arjun Bhargava*, Takaaki Tagawa, Adrien Gaidon\\
Toyota Research Institute (TRI) \\
\{jie.li, allan.raventos, arjun.bhargava, takaaki.tagawa, adrien.gaidon\}@tri.global}

\begin{document}
\maketitle
\begin{abstract}
    We propose an end-to-end learning approach for panoptic segmentation, a novel task unifying instance (things) and semantic (stuff) segmentation. 
    Our model, TASCNet, uses feature maps from a shared backbone network to predict in a single feed-forward pass both things and stuff segmentations. We explicitly constrain these two output distributions through a global things and stuff binary mask to enforce cross-task consistency. Our proposed unified network is competitive with the state of the art on several benchmark datasets for panoptic segmentation as well as on the individual semantic and instance segmentation tasks.

\end{abstract}
\section{Introduction}\label{sec:intro}

Panoptic segmentation is a computer vision task recently proposed by Kirillov et al.~\cite{kirillov2018panoptic}
that aims to unify the tasks of semantic segmentation (assign a semantic class label to each pixel) and instance segmentation (detect and segment each object instance). This task has drawn attention from the computer vision community as a key next step in dense scene understanding ~\cite{kendall2017multi,Li_2018_ECCV,neven2017fast}, and several publicly available benchmark datasets have started to provide labels supporting this task, including Cityscapes~\cite{CordtsCVPR16Cityscapes}, Mapillary Vistas~\cite{neuhold2017mapillary-vistas}, ADE20k~\cite{zhou2017scene}, and COCO~\cite{LinECCV14Microsoft}.

\begin{figure}[ht!]
    \centering
    \subfloat[Semantic segmentation]{\includegraphics[width=0.23\textwidth]{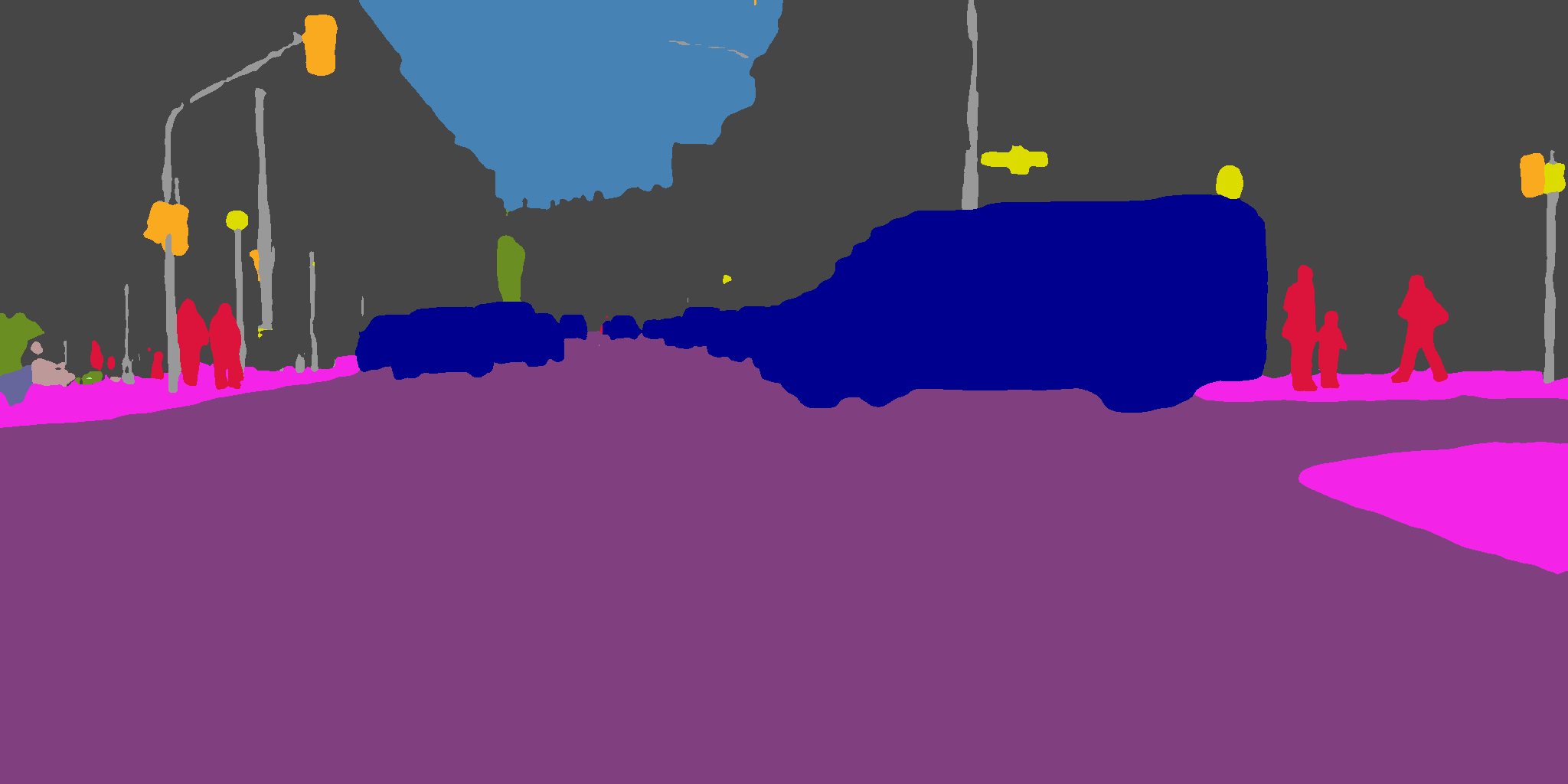}}
    \subfloat[Instance segmentation]{\includegraphics[width=0.23\textwidth]{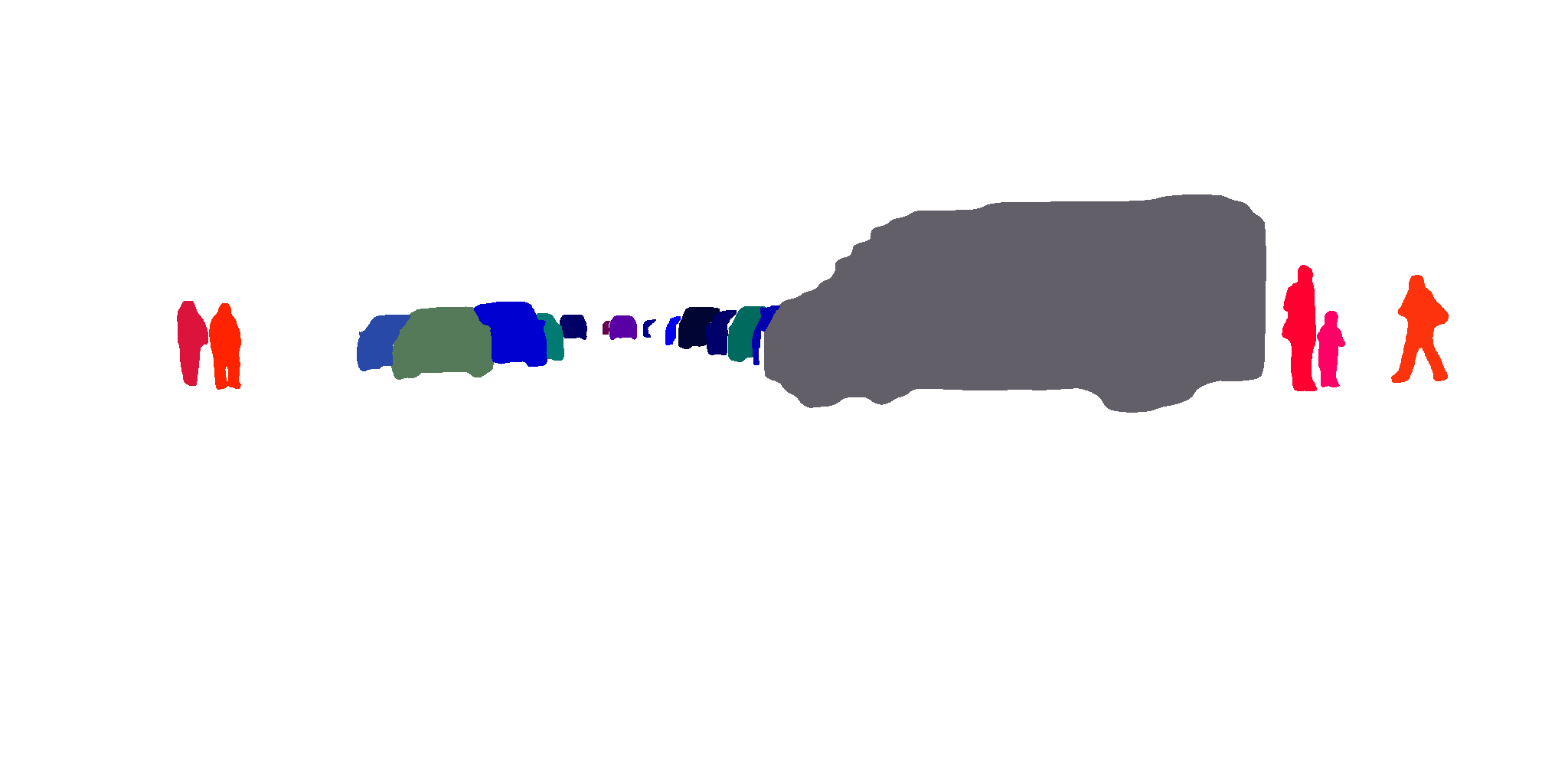}}\\
    \subfloat[Things and Stuff Consistency]{\includegraphics[width=0.23\textwidth]{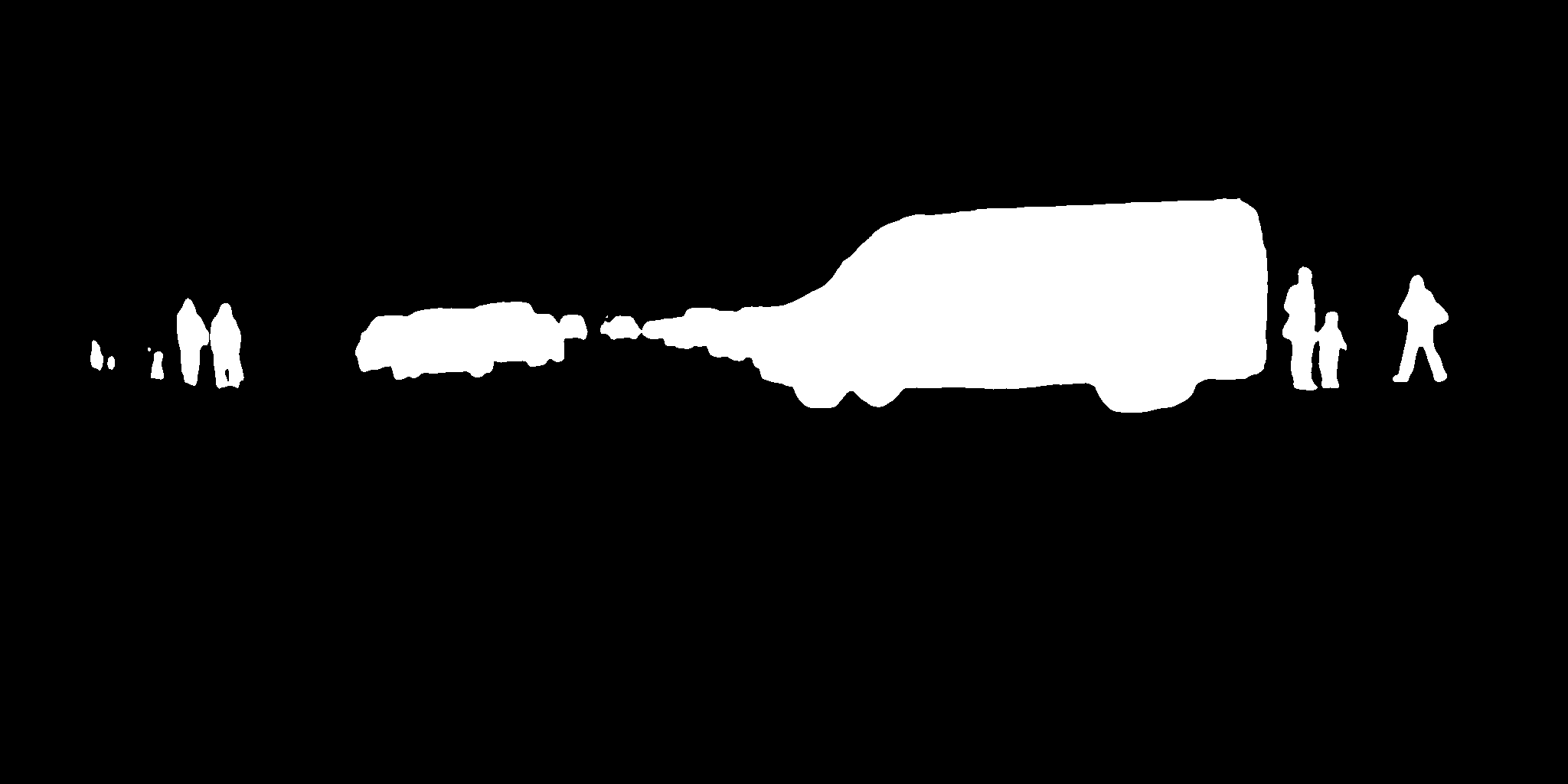}}
    \subfloat[Panoptic segmentation]{\includegraphics[width=0.23\textwidth]{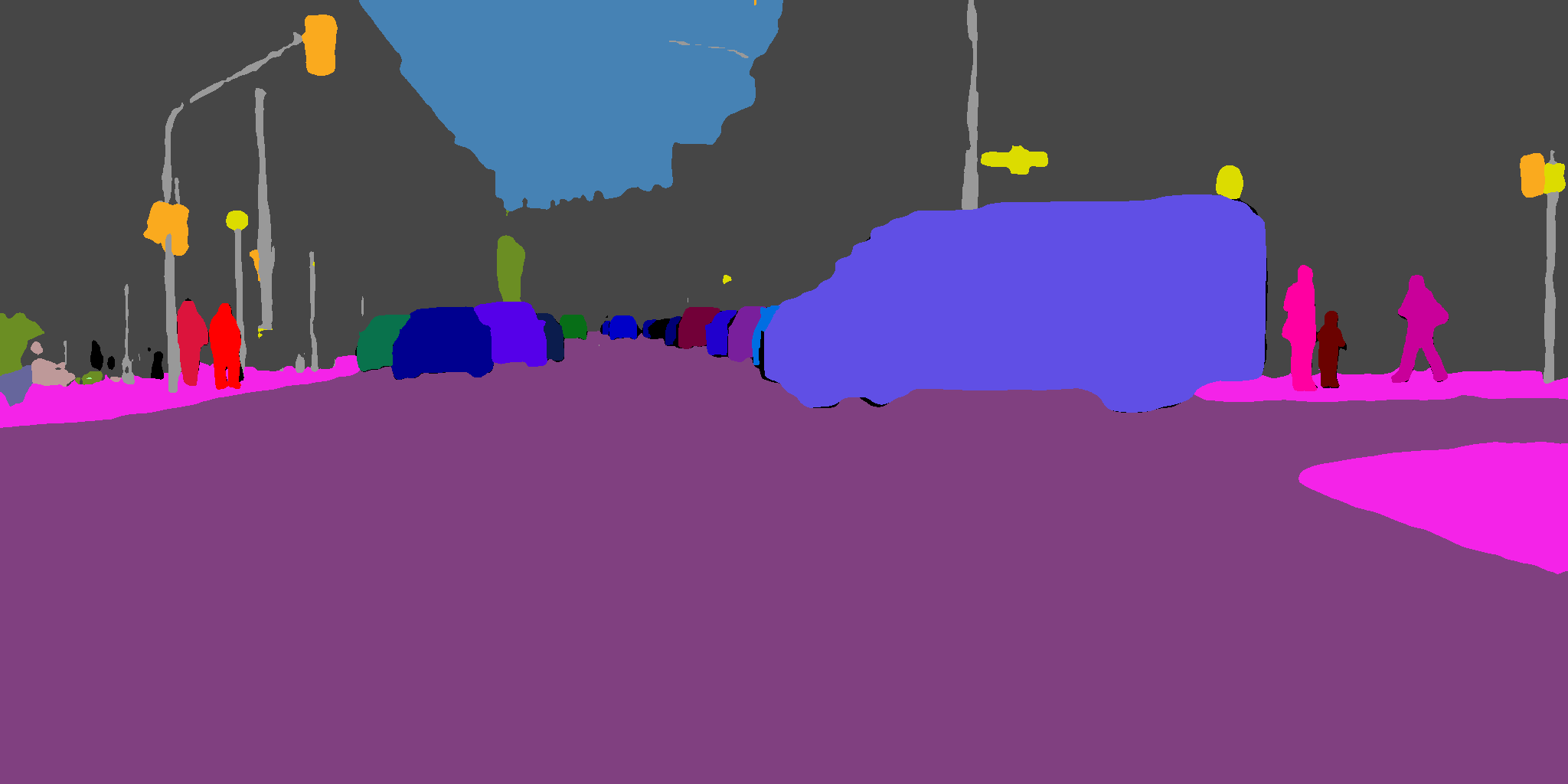}}
    \caption{We propose an end-to-end architecture for panoptic segmentation. Our model predicts things and stuff with a shared backbone and an internal mask enforcing Things and Stuff Consistency (TASC) that can be used to guide fusion.}
    \vspace*{1mm}
    \label{fig:flowchart}
\end{figure}

%
To date, state-of-the-art techniques for semantic and instance segmentation have evolved in different directions that do not seem directly compatible.
On one hand, the best semantic segmentation networks~\cite{bulo2017place, zhao2017pyramid, deeplabv3plus2018} focus on dense tensor-to-tensor classification architectures that excel at recognizing \textit{stuff} categories like roads, buildings, or sky~\cite{adelson2001seeing}. These networks leverage discriminative texture and contextual features, achieving impressive results in a wide variety of scenes.
On the other hand, the best performing instance segmentation methods rely on the recent progress in object detection - they first detect 2D bounding boxes of objects and then perform foreground segmentation on regions of interest~\cite{pinheiro2015learning, he2017mask, liu2018path}. This approach uses the key insight that \textit{things} have a well-defined spatial extent and discriminative appearance features.

The fundamental differences in approaches between handling stuff and things yields a strong natural baseline for panoptic segmentation~\cite{kirillov2018panoptic}: using two independent networks for semantic and instance segmentation followed by heuristic post-processing and late fusion of the two outputs. 

In contrast, we postulate that addressing the two tasks together will result in increased performance for the joint panoptic task as well as for the separate semantic and instance segmentation tasks.

The basis for this hypothesis is the explicit relation between the tasks at the two ends of the modeling pipeline: i) early on at the feature level (capturing general appearance properties), and ii) at the output space level (mutual exclusion, overlap constraints, and contextual relations).

Therefore, the main challenge we address is how to formulate a unified model and optimization scheme where the sub-task commonalities reinforce the learning, while preventing the aforementioned fundamental differences from leading to training instabilities or worse combined performance, a common problem in multi-task learning~\cite{zamir2018taskonomy}.

Our main contribution is a deep network and end-to-end learning method for panoptic segmentation that is able to optimally fuse things and stuff.
Most parameters are shared in a ResNet backbone~\cite{he2016resnet} and a 4-stage \ac{FPN}~\cite{lin2017feature} that is able to learn representations useful for subsequent semantic and instance segmentation heads.
In addition, we propose a new differentiable \ac{TASC} to maintain alignment between the output distributions of the two sub-tasks during training.
This additional objective encourages separation between the outputs of our semantic and instance segmentation heads to be minimal, while simultaneously enabling mask-guided fusion (cf.~Figure~\ref{fig:flowchart}).

Our unified architecture, TASCNet, maintains or improves the performance of individually trained models and is competitive with panoptic quality benchmarks on the Mapillary Vistas~\cite{neuhold2017mapillary-vistas}, Cityscapes datasets~\cite{Cordts2016Cityscapes} and COCO datasets~\cite{lin2014microsoft}.
We conduct a detailed ablative analysis, experimentally confirming that our cross-task constraint is key to improving training stability and accuracy.
Finally, we show that using a single network has the benefit of simplifying training and inference procedures, while improving efficiency by greatly reducing the number of parameters.
%



\section{Related Work}

Tackling dense scene understanding and individual object recognition simultaneously has a long and rich history in computer vision.
Tu et al.~\cite{tu2005image} proposed a hierarchical probabilistic graphical model for scene parsing, disentangling objects, faces, textures, segments, and shapes.
This seminal paper inspired a fertile research direction, including contributions on how to explicitly model the relations between things and stuff categories~\cite{sun2014relating, yao2012describing, tighe2014scene, tighe2013finding, Mottaghi-2014}. For instance, Sun et al.~\cite{sun2014relating} use a CRF over image segments to model geometric and semantic relations. Yao et al.~\cite{yao2012describing} incorporate segmentation unary potentials and object reasoning ones (co-occurrence and detection compatibility) in a holistic structured loss. Tighe et al.~\cite{tighe2013finding} combined semantic segmentation with per-exemplar sliding window.
These approaches rely on handcrafting specific unary and pairwise potentials acting as constraining priors on scene components and their expected relations.

In contrast, deep neural networks can learn powerful shared representations from data, leading to the state of the art in both semantic segmentation~\cite{bulo2017place, zhao2017pyramid, deeplabv3plus2018} and object detection~\cite{ren2015faster, pinheiro2015learning, he2017mask, liu2018path}.
As the corresponding architectures share fundamental similarities inherited from the seminal AlexNet model~\cite{Krizhevsky2012}, several works have naturally leveraged the commonalities to propose multi-task models that can simultaneously address semantic segmentation, object detection, and more~\cite{uhrig2016pixel, kendall2017multi, kokkinos2017ubernet, neven2017fast, teichmann2018multinet}.
These networks typically follow an encoder-decoder architecture, sharing initial layers followed by separate task-specific branches.
In the case of tasks partially competing with each other (e.g., disagreeing on specific image regions), this can result in worse performance (globally and for each task), training instabilities, or outputs not consistent across tasks as noted in~\cite{kirillov2018panoptic}.
In order to better leverage task affinities and reduce the need for supervision, Zamir et al.~\cite{zamir2018taskonomy} build a ``taskonomy" by learning general task transfer functions.
Other works have proposed simple methods tackling the issue of loss weighting. Kendall et al.~\cite{kendall2017multi} propose to use task-dependent uncertainty to weigh the different loss components. Chen et al.~\cite{chen2017gradnorm} propose another weighting based on gradient norms.
Alternatively, Sener et al.~\cite{sener2018multi} formulate multi-task learning as a multi-objective optimization problem.
These approaches avoid complicated hyper-parameter tuning and reduce training times, but only marginally improve joint performance, under-performing larger individual per-task models.


In contrast to these general multi-task learning approaches, we focus explicitly on the relations between stuff and thing categories, with the goal of improving individual performance and addressing the unified panoptic prediction task.
Dai et al.~\cite{dai2015convolutional} predict things and stuff segmentation with a shared feature extractor and convolutional feature masking of region proposals designed initially for object detection. Those are sampled and combined to provide sufficient coverage of the stuff regions, but the relation between things and stuff is not explicitly leveraged.
Chen et al.~\cite{chen2017masklab} leverage semantic segmentation logits to refine instance segmentation masks, but not vice-versa.

Formalizing stuff and things segmentation as a single task, Kirillov et al.~\cite{kirillov2018panoptic} propose a unified metric called \ac{PQ} and a strong late fusion baseline combining separate state-of-the-art networks for instance and semantic segmentation. This method uses a simple non-maximum suppression (NMS) heuristic to overlay instance segmentation predictions on top of a ``background" of dense semantic segmentation predictions.
Saleh et. al~\cite{saleh2018effective} show that this heuristic is particularly effective for sim2real transfer of semantic segmentation by first decoupling things and stuff before late fusion. Indeed, stuff classes can have photo-realistic synthetic textures (ensuring stuff segmentation transfer), while objects typically have realistic shapes (ensuring detection-based instance segmentation generalization). This approach leverages the specificity of things and stuff, but not their relation and does not tackle the joint panoptic task.
Li et al.~\cite{Li_2018_ECCV} propose an end-to-end approach that tackles the unified panoptic problem by reducing it to a semantic segmentation partitioning problem using a fixed object detector and "dummy detections" to capture stuff categories.
Their work focuses on the flexibility to handle weak supervision, at the expense of accuracy, yielding significantly worse performance than the panoptic baseline of~\cite{kirillov2018panoptic}, even when fully supervised.
%



\section{End-to-end Panoptic Segmentation\label{sec:method}}

\subsection{TASCNet Architecture}
\begin{figure*}[!]
    \centering
    \includegraphics[width=\textwidth]{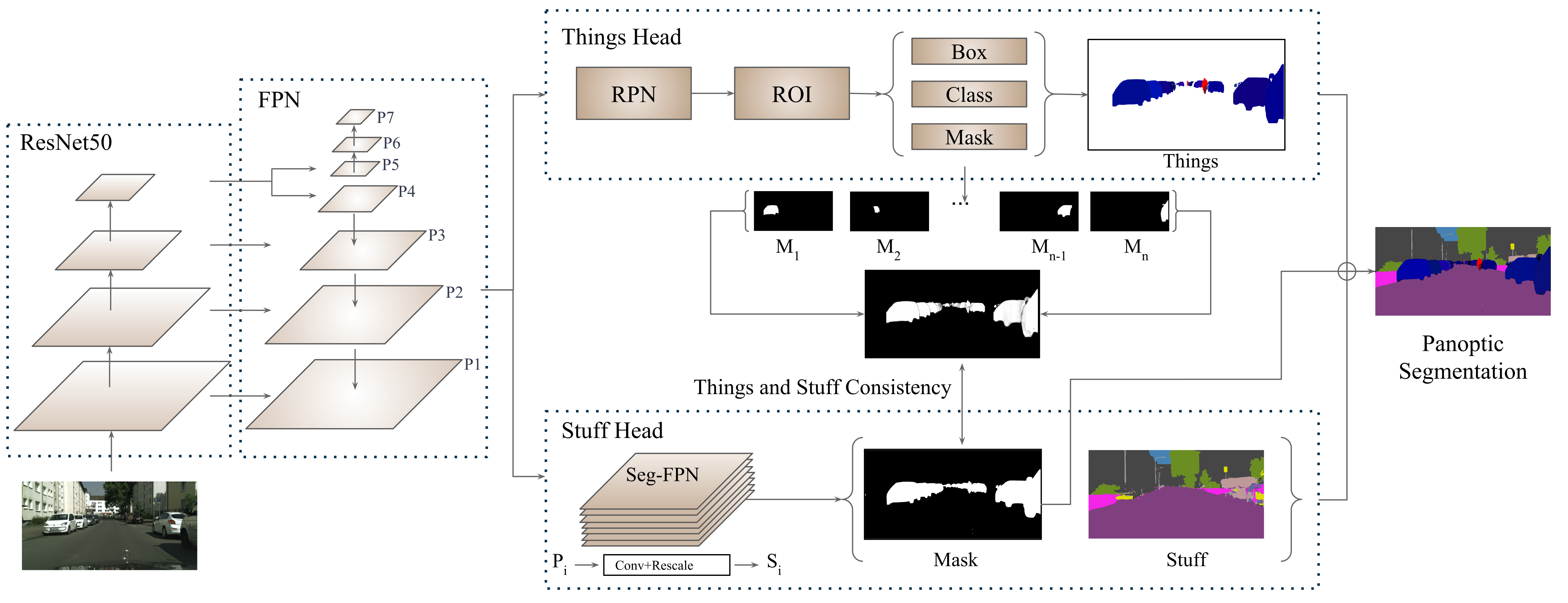}
    \caption{\textbf{TASCNet}: Our unified architecture jointly predicts things, stuff, and a fusion mask. The proposed heads are built on top of a ResNet + FPN backbone. The Stuff Head uses fully convolutional layers to densely predict all stuff classes and an additional things mask. The Things Head uses region-based CNN layers for instance detection and segmentation. In between these two prediction heads, we propose Things and Stuff Consistency loss to ensure alignment between the predictions.}
    \label{fig:joint-network}
\end{figure*}

High-performance models for instance and semantic segmentation share similar structures, typically employing deep backbones that generate rich feature representations on top of which task-specific heads are attached~\cite{deeplabv3plus2018}\cite{he2017mask}. Our TASCNet architecture follows this general motif, as is depicted in Figure~\ref{fig:joint-network}. We use a ResNet50~\cite{he2016resnet} with an \ac{FPN}~\cite{lin2017feature} as our backbone, with two task specific heads that share feature maps from the \ac{FPN}. While the ResNet alone has a large receptive field due to aggressive downsampling, this comes at the expense of spatial resolution and the ability to accurately localize small and large objects. Using an \ac{FPN} enables us to capture low-level features from deeper within the backbone network to recognize a broader range of object scales with far fewer parameters than dilated convolutions. This is a crucial design choice when considering hardware constraints for the \textit{already} memory-intensive semantic and instance segmentation tasks, let alone the joint learning task \cite{lin2017feature}.

\subsubsection{Things Head}
To regress instances, we use Region-based CNN heads on top of the \ac{FPN}, similarly to Mask R-CNN \cite{he2017mask}. We augment the \ac{FPN} with two additional high level context feature maps, similarly to \cite{retinanet}. Improving the instance segmentation architecture was not the primary focus of this work, and we use the same head structures as in \cite{he2017mask}. We train the bounding box regression head, class prediction head, and mask head in an end-to-end fashion.
\subsubsection{Stuff Head\label{subsec:stuff-head}}
Taking inspiration from Kirillov et al~\cite{coco_stuff_2017_fair}, we leverage the multi-scale features from the \ac{FPN} with minimal additional parameters to make dense semantic predictions. From each feature map level of the \ac{FPN}, we:
\begin{enumerate}
    \item apply a set of 3x3 convolutions, reducing the number of channels from 256 to 128;
    \item normalize the layer using a group normalization with 16 groups~\cite{groupnorm};
    \item apply an additional set of 3x3 convolutions, maintaining the number of channels;
    \item normalize and upsample to the largest FPN feature map size (4x downsampled from the input resolution).
\end{enumerate}
Each output layer is then stacked and one final convolution is applied to predict the class per pixel.

In order to provide sufficient information for a full ontology panoptic segmentation in the end, the minimum number of classes to be predicted in the stuff head is $N+1$, which $N$ stuff classes and 1 class for all the things. However, we found that treating all the classes ($N+M$) in the stuff head can help improve the final model performance. More analysis will be provided in Section~\ref{subsec:analysis}
\subsection{Things and Stuff Consistency (TASC)}

Although the sub-task heads are trained using shared features, the output distributions of the two heads can still drift apart. There are several potential causes of this drift, such as minor differences in annotations for instance vs. semantic segmentations, sub-optimal loss functions that capture separate objectives, and local minima for the sub-tasks that do not optimize the joint criterion.

For panoptic segmentation, however, we aim to train towards a global minimum in which the things and stuff segmentations from the two tasks are identical. We seek to enforce such a shared representation through an intermediate confidence mask reflecting which pixels each task considers to be things vs. stuff.

This mask can be constructed in a differentiable manner from both instance and semantic segmentation outputs. Doing so from dense semantic predictions is trivial. First, we apply a threshold of 0.5 to the logits of the Stuff head. Then, all remaining pixels predicting things classes are assigned to their logit values, and all pixels predicting stuff classes are assigned to 0.

For the Things head, constructing the confidence mask is slightly more involved. At train time, an \ac{RPN} proposes regions of interest (RoI), which are pooled to a fixed size using an RoI-Align operation. The mask head then produces a per-class foreground/background confidence mask for each positive  proposal from the \ac{RPN}. We can then re-assemble the global binary mask using an operation we dub ``RoI-Flatten": 

\begin{enumerate}
    \item For each image, we construct an empty tensor of equivalent size to the input image.
    \item For each RoI, we only consider the foreground/background mask for the class of the ground truth instance the RoI was assigned to regress, 
    \item We interpolate each of these single-instance masks, ($M_1$, ..., $M_K$), to the size of its corresponding RoI in the input image.
    \item A threshold of $0.5$ is applied to each mask. The thresholded mask is then added to the RoI's original position in the empty tensor.
    \item To obtain our final confidence mask, we normalize by the instance count at each pixel post-threshold.
\end{enumerate}

\begin{figure}
    \centering
    \includegraphics[width=0.22\textwidth]{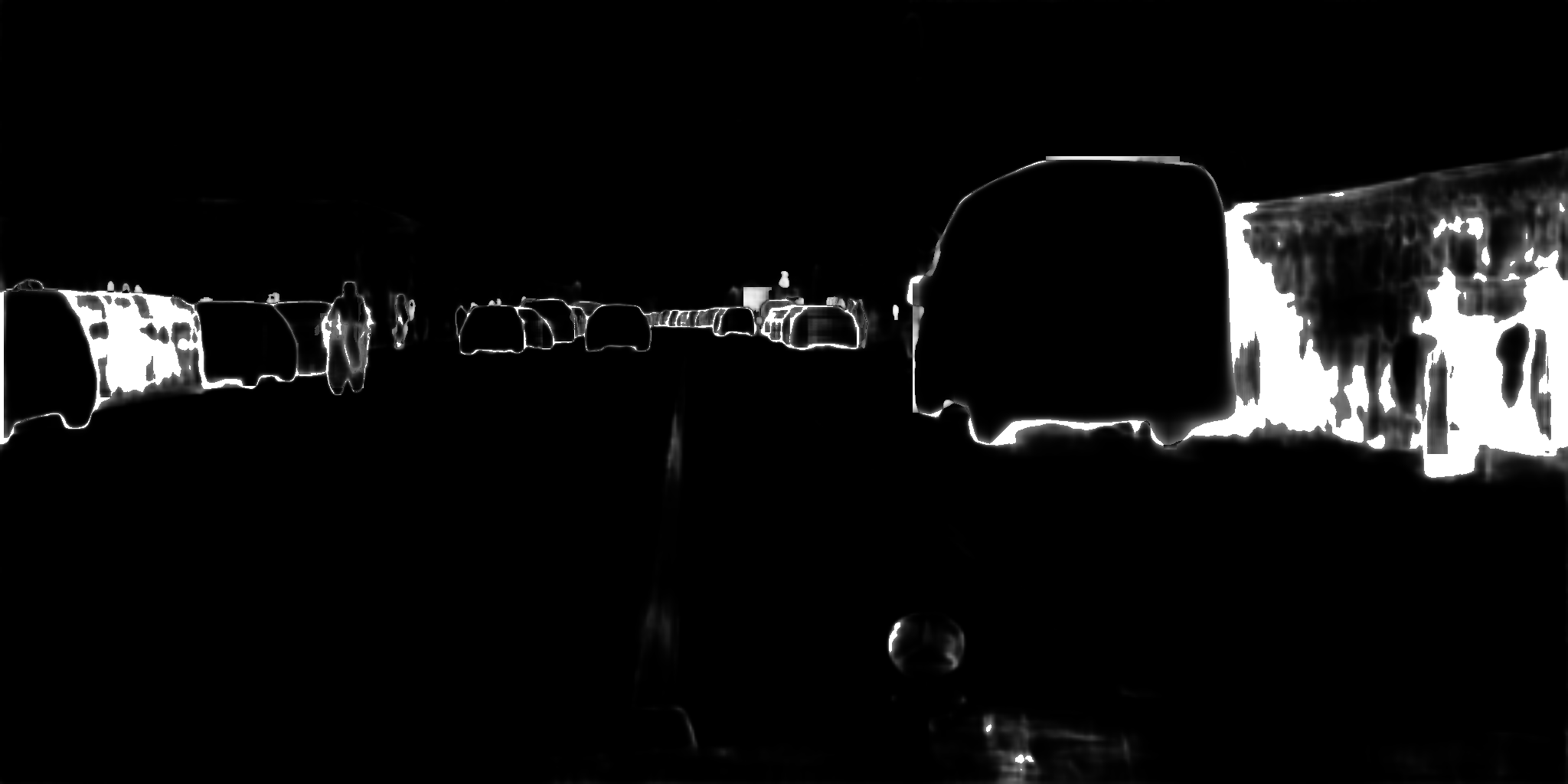}
    \includegraphics[width=0.22\textwidth]{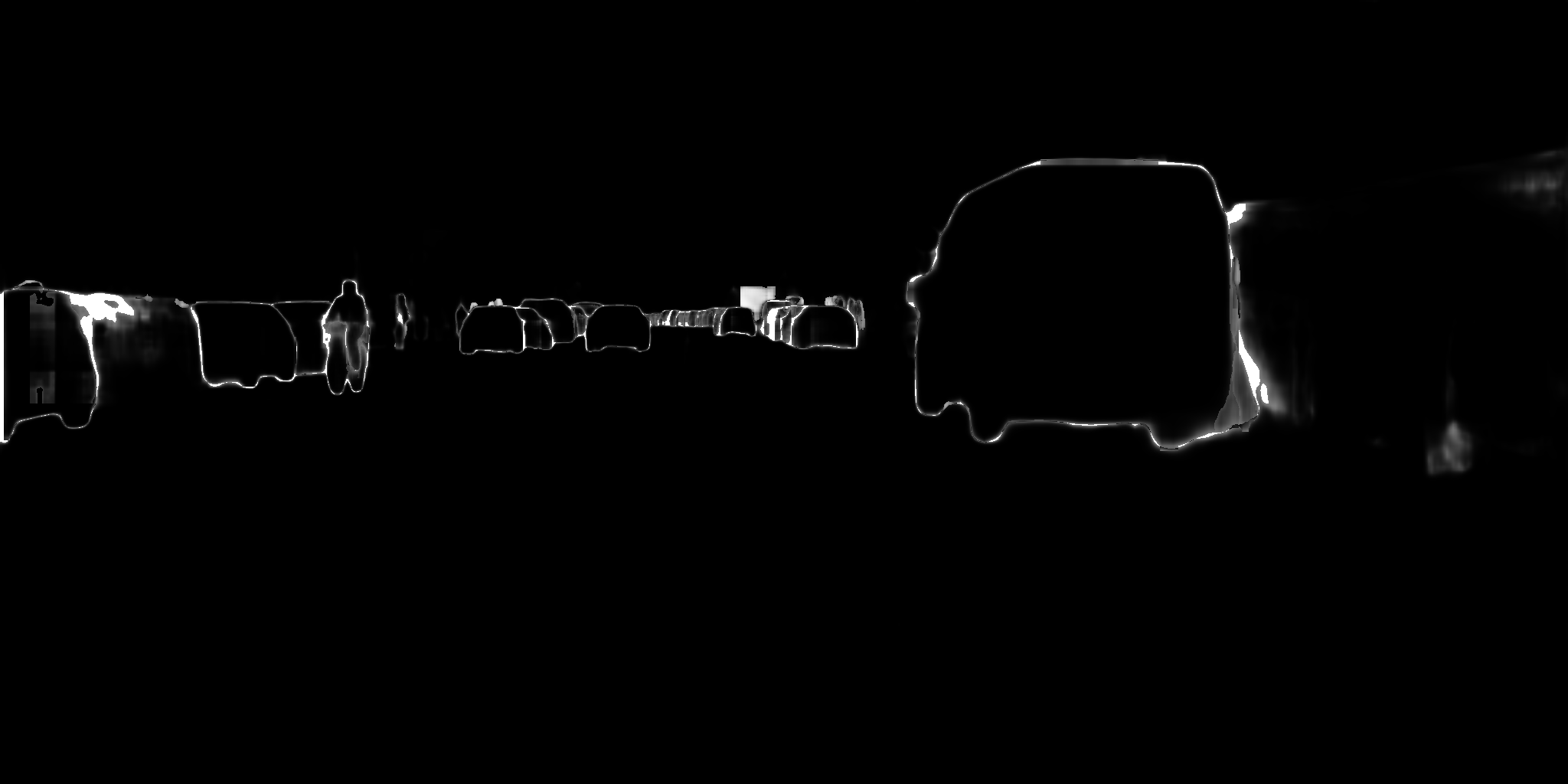}
    \caption{\textbf{Residual Example}. Example image of residuals from a model trained without TASC (left) and a model trained with TASC (right).}
    \label{fig:residual}
\end{figure}

To encourage our instance and semantic segmentation heads to agree on which pixels are things and which are stuff, we minimize the residual between these two masks using an $L_2$ loss. This residual is visualized in Figure ~\ref{fig:residual}.

\begin{figure}
    \centering
\includegraphics[width=0.3\textwidth]{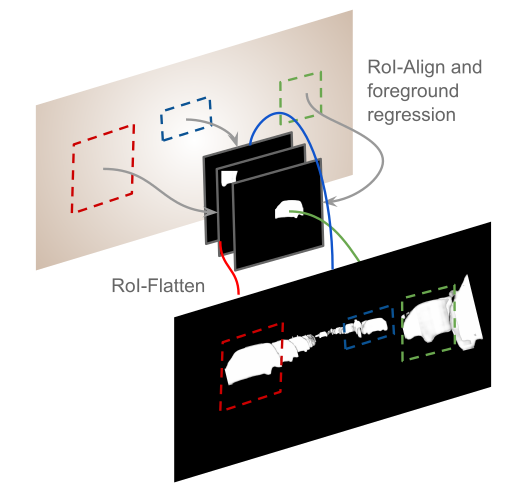} \hfill \\
    \vspace{4mm}
    \caption{\textbf{RoI-Flatten}. We proposed a differentiable operation to merge individual proposal masks into a binary mask to provide global constrain across tasks.}
    \vspace*{1mm}
    \label{fig:roi}
\end{figure}

\subsection{Mask-Guided Fusion}

Our learning objective encourages the two masks to agree. Therefore, in a converged TASCNet we can use the semantic segmentation mask to select which pixels are obtained from the instance segmentation output and which pixels are obtained from the semantic segmentation output. 

We consequently define a simple post-processing procedure: we add regressed instances into the final panoptic output in decreasing order of confidence, only adding an instance to the output if it has an IoU of under 0.4 with instances that have already been added and an IoU of greater than 0.7 with the mask.

\section{Experiments}\label{sec:exp}
\subsection{Datasets}

We evaluate our proposed approach using different benchmark datasets, Cityscapes~\cite{Cordts2016Cityscapes}, Mapillary Vistas~\cite{neuhold2017mapillary-vistas} and COCO~\cite{lin2014microsoft}.
These datasets have large gaps in both complexity and image content.

Cityscapes is comprised of street imagery from Europe and has a total of 5000 densely annotated images with an ontology of 19 classes, 8 thing classes and 11 stuff classes. All the images are at 1024 x 2048 resolution, and are split into separate training, validation and test sets. We train our model on the ``fine" annotations in the training set and test on the provided validation set.

The Mapillary Vistas dataset, on the otherhand, while still a street scene dataset, is far more challenging. It consists of a wide variety of geographic settings, camera types, weather conditions, image aspect ratios, and object frequencies. The average resolution of the images is around $9$ megapixels, which causes considerable complications for training deep nets with limited memory. The dataset consists of $18000$ training images, $2000$ validation images, and $5000$ testing images. Annotations for pixel-wise semantic segmentation and instance segmentation are available for the training and validation sets. The labels are defined on an ontology of $65$ semantic classes, including $37$ thing classes and $28$ stuff classes. %

{COCO} is a large scale object detection and segmentation dataset. We use its 2017 edition with $118k$ training images, $5k$ validation images and $2k$ testing images. The labels consist of 133 classes. $80$ classes have instance level annotations.

\subsubsection*{Evaluation Metrics}
     We evaluate our model's performance on Panoptic Segmentation task using the \ac{PQ} metric proposed by~\cite{kirillov2018panoptic}, %
     \begin{equation}
      PQ = \frac{\Sigma_{(p,g)\in TP}{IoU}_{(p,g)}}{|TP|+\frac{1}{2}|FP| + \frac{1}{2}|FN|}
    \end{equation}
    where \textit{p} and \textit{g} are matched predicted and ground truth segments exceeding an IoU threshold of $0.5$, and TP, FP, FN denote true positives, false positives, and false negatives, respectively.
    
    To better explore the capability of our proposed approach, we also evaluate our model performance on the task of instance segmentation and semantic segmentation. 
    For semantic segmentation, we use the standard metric, PASCAL VOC \ac{IoU}.
    
    For instance segmentation, following recent literature, we average over the $AP^r$~\cite{hariharan2014simultaneous} with acceptance \ac{IoU} from $0.5$ to $0.95$ in increments of $0.05$. We call this metric $AP$ in the rest of the paper without ambiguity. For minor scores, we also report the $AP50=AP^r(IoU>0.5)$.



\subsection{Implementation Details}
We conduct most of our experiments on P3.16xlarge instances on Amazon Web Services (AWS), each of them features 8 V100 GPUs. 

For Cityscapes experiments, we train our models for $20000$ iterations with 1 image per GPU, $0.01$ base learning rate, 0.0001 weight decay. We divide the learning rate by a factor of 10 at iteration 14000 and 18000.
For Mapillary Vistas experiments, we follow the same settings for Cityscapes except for a longer schedule to account for the size of dataset. We train the models for $180000$ iterations and decrease the learning rate at $120000$ and $16000$. In this set of experiments, due to the limitation of GPU memory, we collapes the prediction ontology in stuff head from $N+M$ to $N+1$ merging all the $M$ things classes as a single ``thing'' class. Therefore, a full ontology $mIoU$ is not available.
For COCO experiments, we follows the 1x training schedule used in~\cite{he2017mask}. We train our models for $90000$ iterations with 2 images per GPU, $0.02$ base learning rate and $0.0001$ weight decay.

For data augmentation, we apply aspect ratio-preserving scale with jitter and randomized left-right flip. In Cityscapes and Vistas experiments, we randomly sample a shortest side from $800~1400$. We apply a longest side maximum of $2500$ to Vistas dataset due to its wide range of image sizes. In COCO experiments, we randomly sample a shortest side from $[640, 800]$ with a maximum longest side of $1333$.

During inference, we apply per-class NMS with a cutoff threshold of $0.3$ for bounding boxes proposals and another run of mask level NMS with the same threshold over all the proposals. For test time augmentation, we conduct inference at multiple scales, each with a horizontal flip. 
\begin{table}[ht!]
\centering
\resizebox{0.49\textwidth}{!}{%
\begin{tabular}{|l|l|l|l|l|l|l|l|}
\hline
Method & Backbone & PQ & PQ th. & PQ st. & mIoU & AP \\ \hline
Kirillov et al~\cite{kirillov2018panoptic} & R50+X101 & 61.2 & 54.0 & 66.4 & N/A & 36.4 \\ \hline\hline
Li at el~\cite{li2018weakly} & R-101 & 53.8 & 42.5 & 62.1 &\textbf{79.8} & N/A \\ \hline
TASCNet& R-50  & {59.3} & \textbf{56.3} & {61.5} & {78.1} & {37.6} \\ \hline
TASCNet(M.)& R-50  & \textbf{60.4} & {56.1} & \textbf{63.3} & {78.7} & \textbf{39.09} \\ \hline
\end{tabular}%
}
\caption{\textbf{Cityscapes Panoptic Segmentation Results}. Our joint network with a small encoder backbone (ResNet-50) is comparable to other state of the art techniques using separate models.\label{tab:cityscapes}}
\end{table}
\begin{table}[ht!]
\centering
\resizebox{0.49\textwidth}{!}{%
\begin{tabular}{|l|l|l|l|l|l|l|l|}
\hline
Method & Backbone & PQ & PQ th. & PQ st. & AP \\ \hline
MaskRCNN+FPN Seg.&2xR-50  & {32.6} & {31.8} & {33.7} & {20.2} \\ \hline\hline
TASCNet& R-50  & {32.6} & {31.1} & {34.4} & {18.5} \\ \hline
TASCNet(M.)& R-50  & \textbf{34.3} & {\textbf{34.8}} & {33.6} & \textbf{20.4} \\ \hline

\end{tabular}%
}
\caption{\textbf{Vistas Panoptic Segmentation Results}. We compare our segmentation performance of the unified network to separate networks for instance and semantic segmentation.\label{tab:Vistas}}
\end{table}
\begin{table*}{}
\centering
 \resizebox{0.7\textwidth}{!}{%
    \begin{tabular}{cccccccccc}
    \hline
    \multicolumn{1}{|l|}{Method Spec.} & \multicolumn{1}{l|}{PQ}  & \multicolumn{1}{l|}{SQ}& \multicolumn{1}{l|}{RQ}& \multicolumn{1}{l|}{PQ th.}& \multicolumn{1}{l|}{SQ th.}& \multicolumn{1}{l|}{RQ th.} & \multicolumn{1}{l|}{PQ st.} &\multicolumn{1}{l|}{SQ st.}&\multicolumn{1}{l|}{RQ st.} \\ \hline

    \multicolumn{1}{|l|}{MMAP-seg} & \multicolumn{1}{l|}{32.2} & \multicolumn{1}{l|}{76.0} & \multicolumn{1}{l|}{40.8} & \multicolumn{1}{l|}{39.0} & \multicolumn{1}{l|}{78.2} & \multicolumn{1}{l|}{49.1} & \multicolumn{1}{l|}{22.0} & \multicolumn{1}{l|}{72.8} & \multicolumn{1}{l|}{28.4}  \\ \hline
    
    \multicolumn{1}{|l|}{MPS-TU} &
    \multicolumn{1}{l|}{27.2} & \multicolumn{1}{l|}{71.9} & \multicolumn{1}{l|}{35.9}  & 
    \multicolumn{1}{l|}{29.6} & \multicolumn{1}{l|}{71.6} & \multicolumn{1}{l|}{39.4} & 
    \multicolumn{1}{l|}{23.4} & \multicolumn{1}{l|}{72.3} & \multicolumn{1}{l|}{30.6}\\ \hline
    
    \multicolumn{1}{|l|}{LeChen} & 
    \multicolumn{1}{l|}{26.2} & \multicolumn{1}{l|}{74.2} & \multicolumn{1}{l|}{33.2} & 
    \multicolumn{1}{l|}{31.3} & \multicolumn{1}{l|}{76.2} & \multicolumn{1}{l|}{39.3} & 
    \multicolumn{1}{l|}{18.7} & \multicolumn{1}{l|}{71.2} & \multicolumn{1}{l|}{24.1} \\ \hline
    
     \multicolumn{1}{|l|}{TASCNet} & \multicolumn{1}{l|}{\textbf{40.7}} & \multicolumn{1}{l|}{\textbf{78.5}}& \multicolumn{1}{l|}{\textbf{50.1}}& \multicolumn{1}{l|}{\textbf{47.0}}& \multicolumn{1}{l|}{\textbf{80.6}}& \multicolumn{1}{l|}{\textbf{57.1}} & \multicolumn{1}{l|}{\textbf{31.0}}& \multicolumn{1}{l|}{\textbf{75.3}}& \multicolumn{1}{l|}{\textbf{39.6}}  \\ \hline
    \end{tabular}%
    }
\caption{\textbf{COCO (Test-dev) Panoptic Segmentation Results.} We report TASCNet single model performance on COCO test-dev server without any test-time augmentation. We compare our model to the challenge entries on the panoptic segmentation leaderboard that also uses unified networks.}
\label{tab:coco}
\end{table*}
\subsection{Experimental Results}
We compare our approach with other state-of-the-art panoptic segmentation models and state-of-the-art models for instance and semantic segmentation on the $3$ datasets. In Table~\ref{tab:cityscapes}, Table~\ref{tab:Vistas} and Table~\ref{tab:coco}, we present this comparison over the relevant metrics for each task. All experiments are carried out using a ResNet-50 (R50) backbone.

We compare to the challenging baseline proposed in~\cite{kirillov2018panoptic}, which combines the state-of-the-art models from single task semantic segmentation (PSPNet~\cite{zhao2017pyramid}) and instance segmentation (Mask R-CNN~\cite{he2017mask}). The panoptic segmentation performance of our proposed TASCNet, using only ResNet-50 backbone and basic test time augmentation, matches the $PQ$ reported in~\cite{kirillov2018panoptic}.
We also compared to Li et. al.~\cite{Li_2018_ECCV}, who also presented their model performance in fully supervised scenario. The panoptic segmentation performance of our unified TASCNet outperforms their fully supervised joint solution~\cite{Li_2018_ECCV} by a large margin. 
It is also worth noting that our single task semantic segmentation outputs generated from panoptic segmentation results is comparable to the state-of-the-art segmentation only methods using very large backbones such as~\cite{bulo2017place}. 



Limited previous panoptic segmentation performance has been reported on the Mapillary Vistas dataset. \cite{kirillov2018panoptic} reported a reference $PQ$ of $38.3$ on a subset of the Vistas test set, combining the winning entries from the in LSUN'17 Segmentation Challenge~\cite{zhao2017pyramid,liulsun}. Megvii report their ECCV'18 Panoptic Segmentation Challenge winning entry in~\cite{megvii-challenge}. However, it is difficult to make a fair comparison to their models without additional technical details. Instead, we treat our own separate modules as baselines, generating the panoptic output as in \cite{kirillov2018panoptic}. We also present our network performance under different settings. 

In Table~\ref{tab:coco}, Our TASCNet outperform other unified-model entries on COCO Panoptic Segmentation Leaderboard. In our prediction, no test-time augmentation or extra heuristic post-processing for COCO, which typically used in challenge entries~\cite{megvii-challenge}, e.g. threshold small stuff area, heuristically not suppress ties on top of human beings, and etc.
\begin{table}[ht!]
\centering
\resizebox{0.49\textwidth}{!}{%
\begin{tabular}{|l|l|l|l|l|l|l|l|}
\hline
Stuff Onto. & $\lambda$ & M. & PQ & PQ th. & PQ st. & mIoU & AP \\ \hline
N+1&1.0&&58.6&54.5&61.5&-&37.7\\\hline
N+M&$0.0$(No TASC)&  & {58.5} & {54.8} & {61.3} & {76.5}& {37.3}\\ \hline
N+M&$0.1$&& {59.3} & {56.3} & {61.5} & {78.1} & {37.6} \\ \hline
N+M&$0.5$& &{59.2} & {55.9} & {61.5} & {78.1} & {37.3} \\ \hline
N+M&$1.0$& &{59.0} & {55.6} & {61.5} & {77.6} & {37.6} \\ \hline
N+M&$1.0$&\cmark & {60.4} & {56.1} & {63.3} & {78.7} & {39.09} \\ \hline

\end{tabular}%
}
\caption{\textbf{Ablation analysis on important network components}. We conduct ablation analysis on different components of the proposed method. $M$ and $N$ indicates the number of things classes and stuff classes respectively. $\lambda$ indicates the weight of TASC loss. `M.' indicates the use of test time augmentation. The final TASCNet configuration is included in the last row.}
\label{tab:analysis}
\end{table}




\begin{table}[ht!]
\centering
\resizebox{0.48\textwidth}{!}{%
\begin{tabular}{|l|l|l|l|l|}
\hline
{Pretrain Data} &{Training Strategy}& PQ & {PQ th.} & {PQ st.} \\ \hline
ImageNet& Joined & 55.9 & 50.5 & 59.8 \\\hline
COCO& Stuff Head + Joined & 57.3 & 51.6 & 61.5 \\\hline
COCO& Things Head + Joined & 58.5 & 55.0 & 61.0 \\ \hline
COCO& Joined& \textbf{59.2} & \textbf{56.0} & \textbf{61.5} \\\hline
\end{tabular}%
}
\caption{\textbf{TASCNet training strategies on Cityscapes.} We compare our approach given different pretrained backbones. We also compare different training strategy (Single pass joined training vs. stage-wise training).}
\label{tab:training-protocol}
\end{table}
\subsection{Ablative Analysis\label{subsec:analysis}}
In this section, we present a thorough ablation study of our proposed method on the Cityscapes dataset. We compare key hyper-parameters as well as variations to important components of the network. We also compare different training strategies on the model.

In Table~\ref{tab:analysis}, we compare different variation to the proposed model on key components. Let $\lambda$ denotes the weight to TASC loss. We compare the model performance under different $\lambda$ values as well as $\lambda=0$, which indicates a model without the propose TASC loss. The weight for all the other losses are $1.0$ everywhere else. The experimental results indicates that the use of TASC loss improves the model performance in PQ with a small variation base on the weight. We choose $\lambda=1$ as our final configuration in all of our other experiments. We also make a comparison on the prediction ontology of stuff head. As discussed Section~\ref{subsec:stuff-head}, the minimum number of classes to be predicted in stuff head is $N+1$ with $N$ stuff classes and a collapsed 'things' class. We find that predicting a full ontology (N stuff classes and M thing classes) in the stuff head help improve the final performance.





For the second part of the analysis, we explore the training protocol for our joint network. We pretrain on various datasets and also examine stage-wise training, as shown in Table ~\ref{tab:training-protocol}. 
It is obvious that a backbone pretrained from COCO datasets in general helps to improve the final panoptic segmentation on Cityscapes.
In the exploration of stage-wise training, we start with training our model with only one head attached, things head or stuff head. When the single-head model converged, we attached the other head and the TASC, then retrain the model. The experimental results indicates that joint training without any fully trained head tends to converge to better minima. 

Please refer to the appendix for more detailed results.

\begin{figure*}[htb!]
\setlength{\tempwidth}{.32\linewidth}
\setlength{\tempheight}{0.16\linewidth}
\centering
\columnname{\textbf{Cityscapes}}\\
\subfloat{\includegraphics[width=\tempwidth,height=\tempheight]{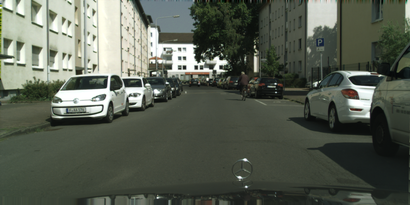}}\hfil
\subfloat{\includegraphics[width=\tempwidth,height=\tempheight]{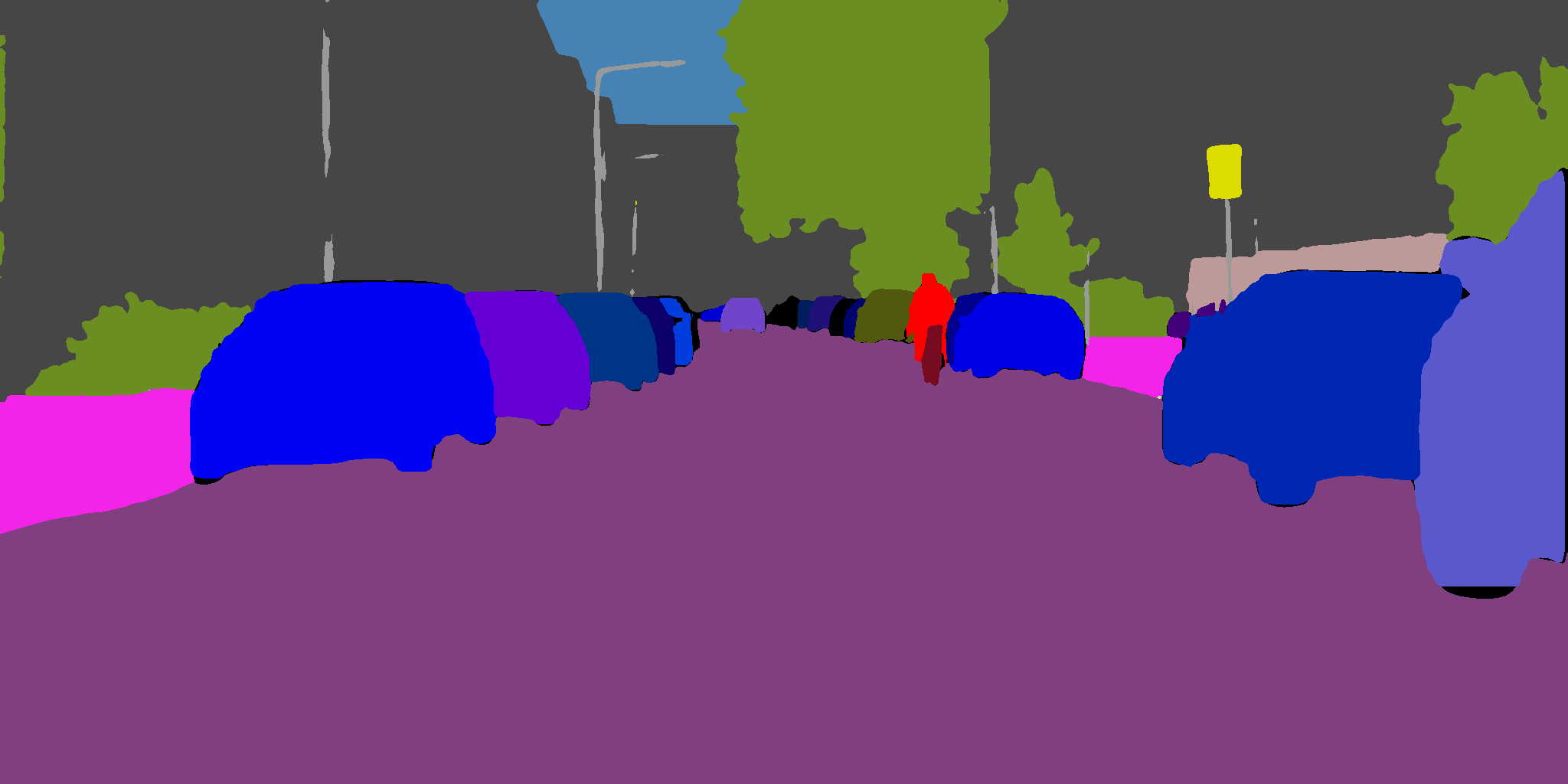}}\hfil
\subfloat{\frame{\includegraphics[width=\tempwidth,height=\tempheight]{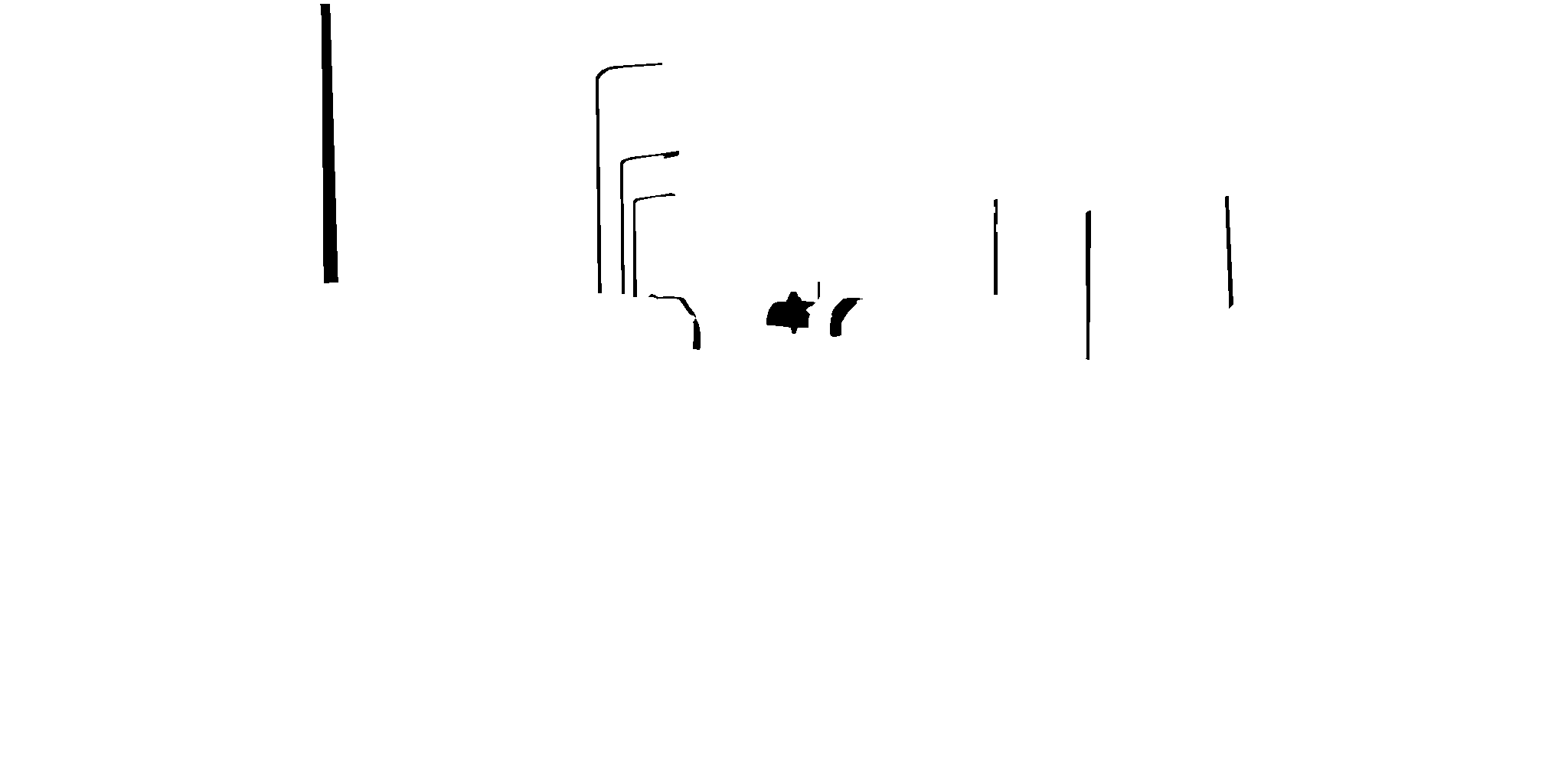}}}\\
\subfloat{\includegraphics[width=\tempwidth,height=\tempheight]{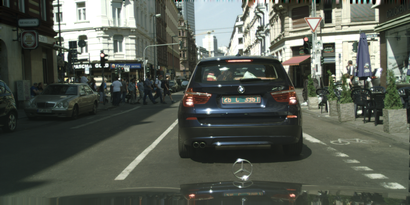}}\hfil
\subfloat{\includegraphics[width=\tempwidth,height=\tempheight]{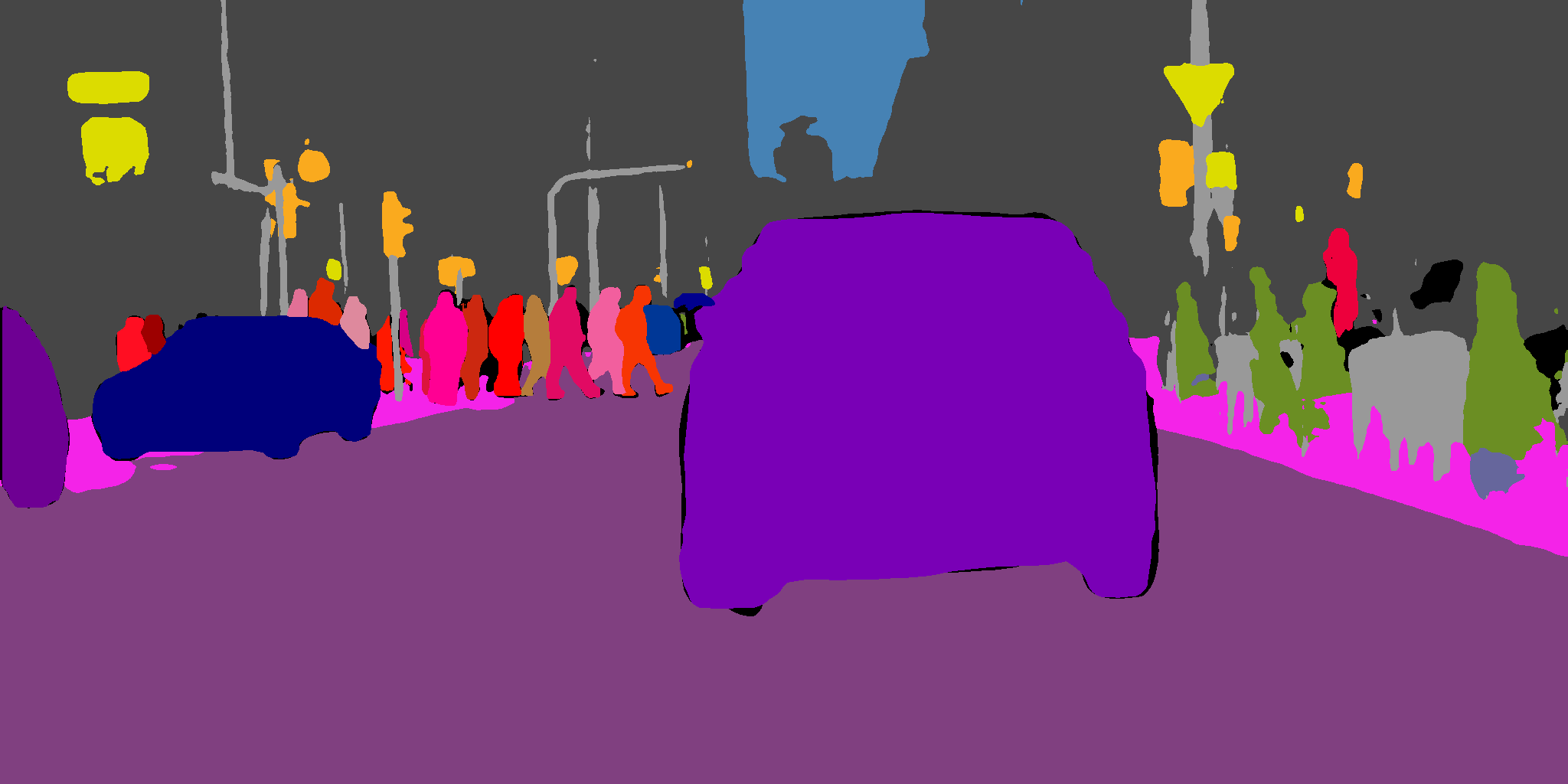}}\hfil
\subfloat{\frame{\includegraphics[width=\tempwidth,height=\tempheight]{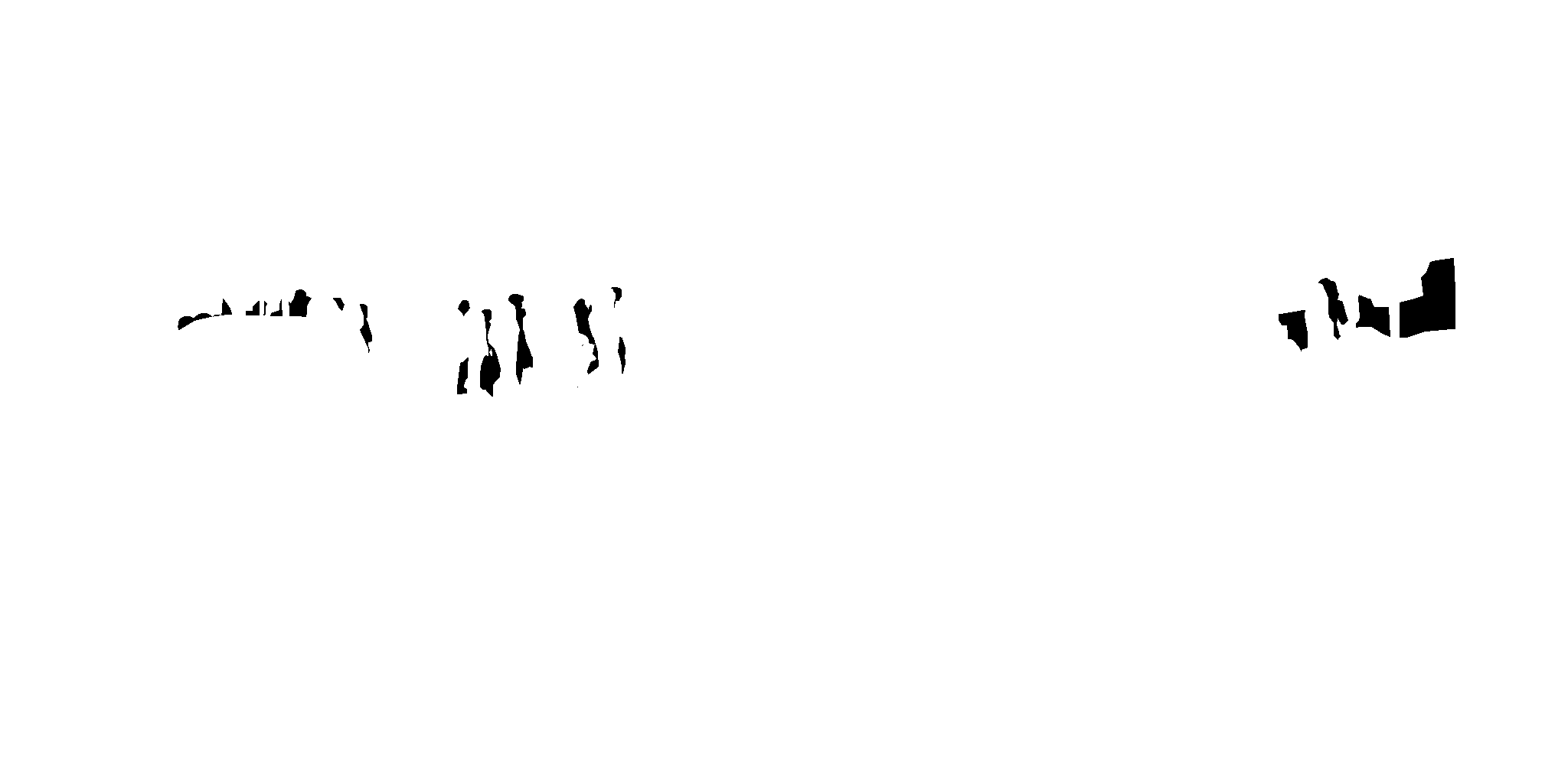}}}\\
\subfloat{\includegraphics[width=\tempwidth,height=\tempheight]{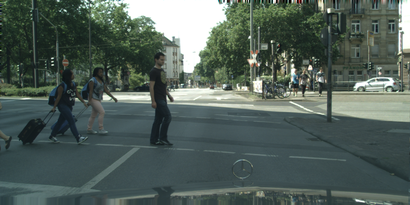}}\hfil
\subfloat{\includegraphics[width=\tempwidth,height=\tempheight]{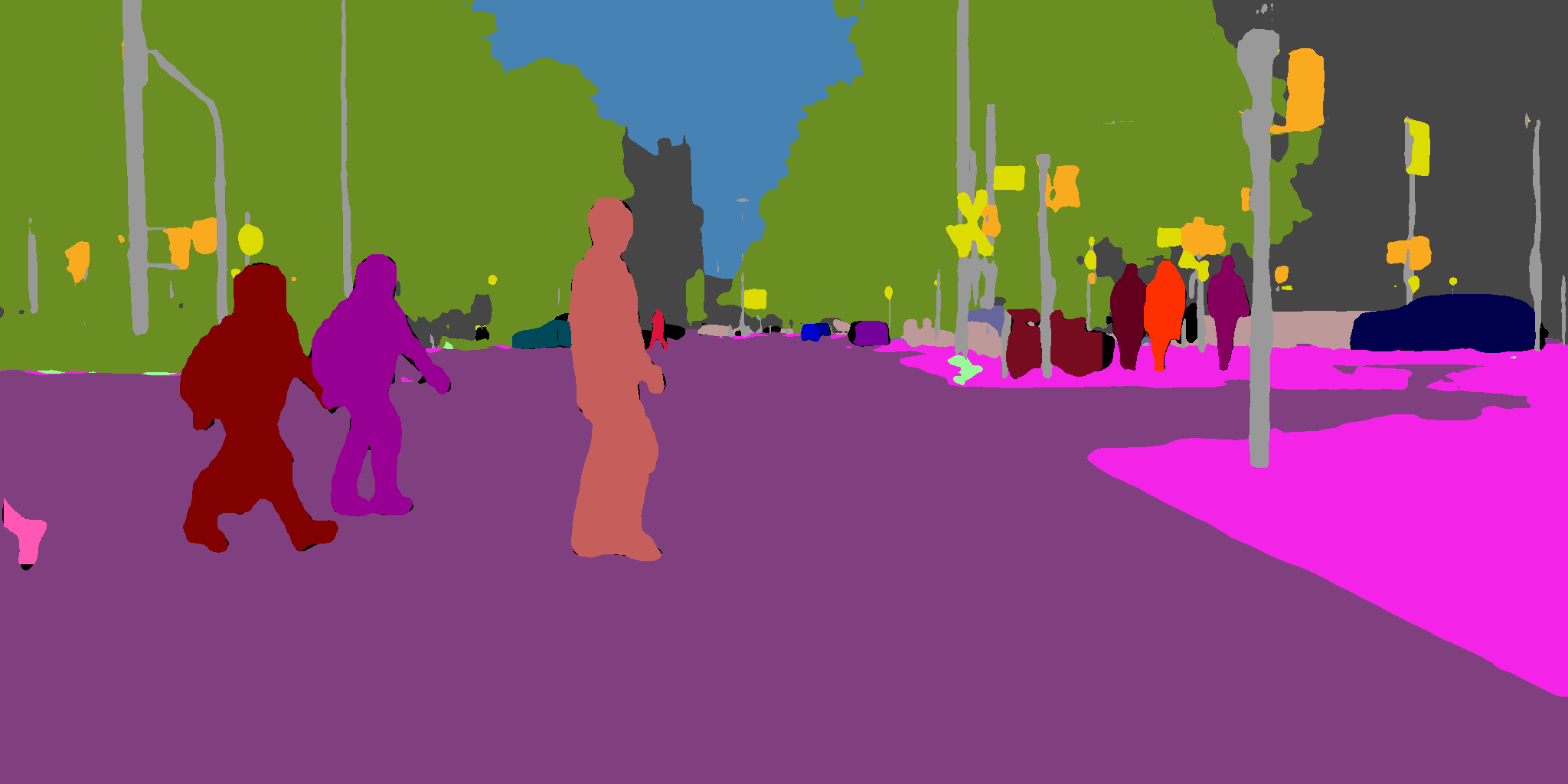}}\hfil
\subfloat{\frame{\includegraphics[width=\tempwidth,height=\tempheight]{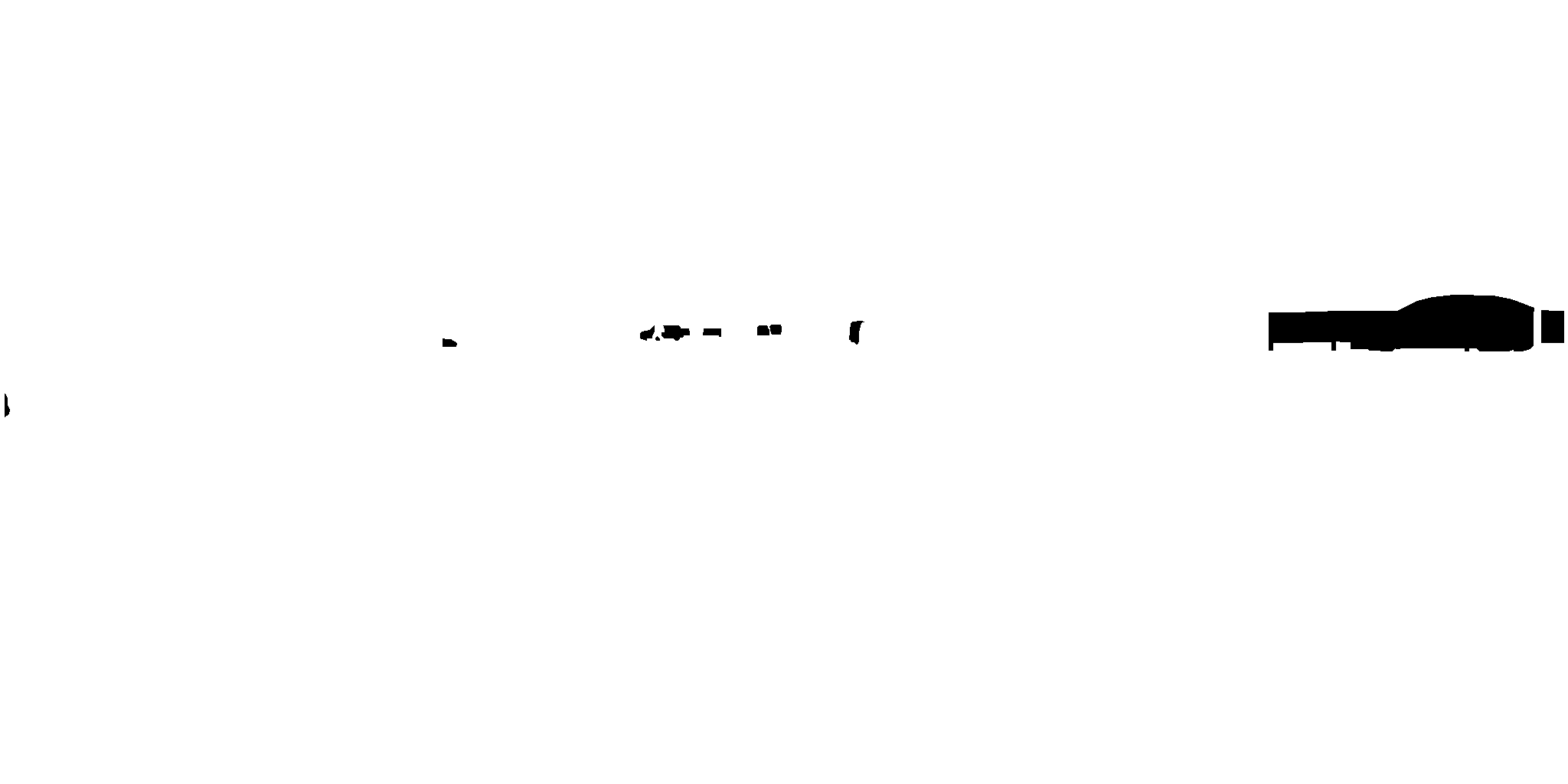}}}\\
\subfloat{\includegraphics[width=\tempwidth,height=\tempheight]{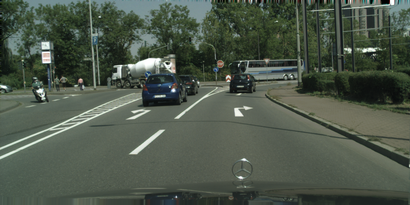}}\hfil
\subfloat{\includegraphics[width=\tempwidth,height=\tempheight]{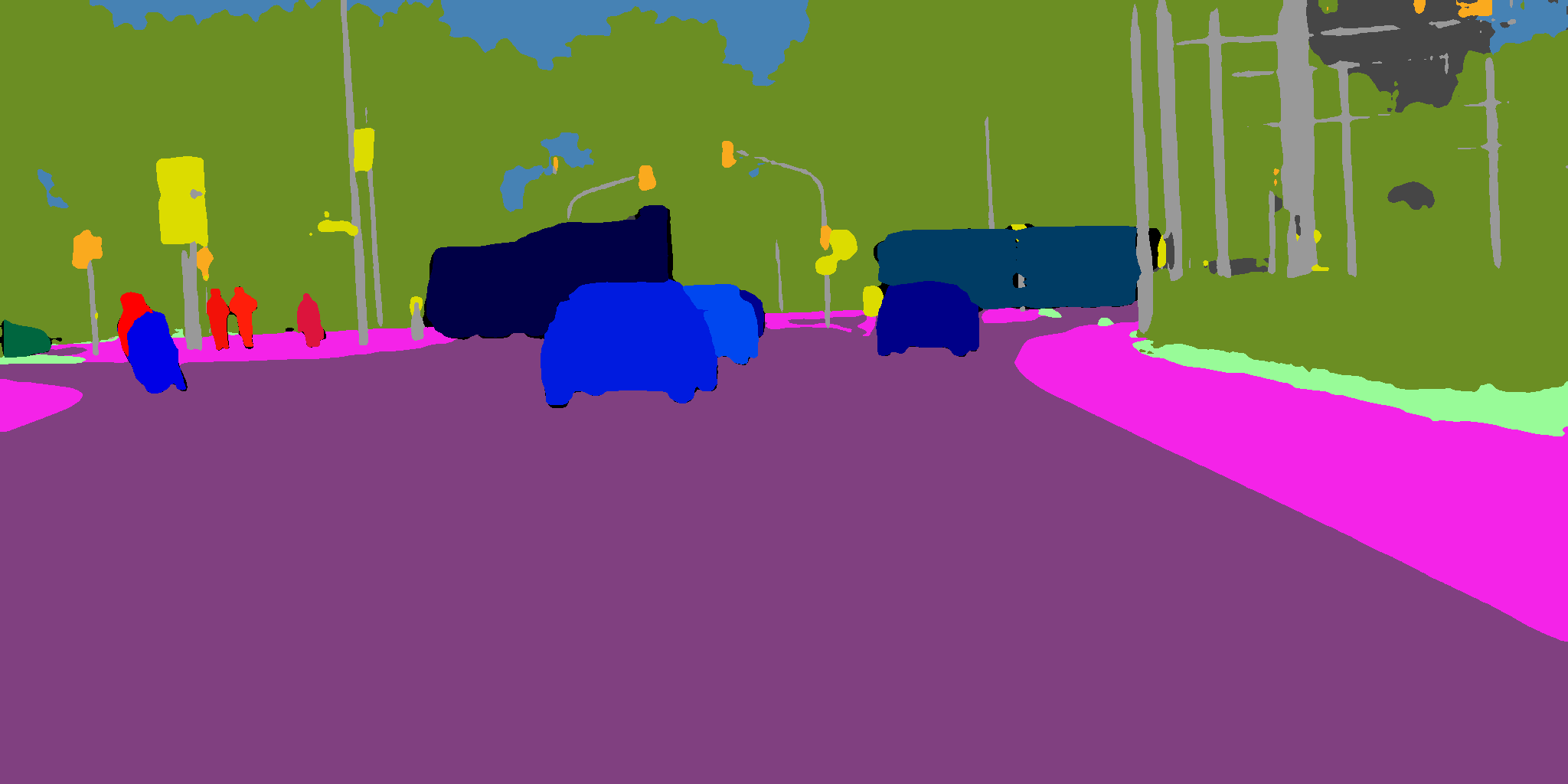}}\hfil
\subfloat{\frame{\includegraphics[width=\tempwidth,height=\tempheight]{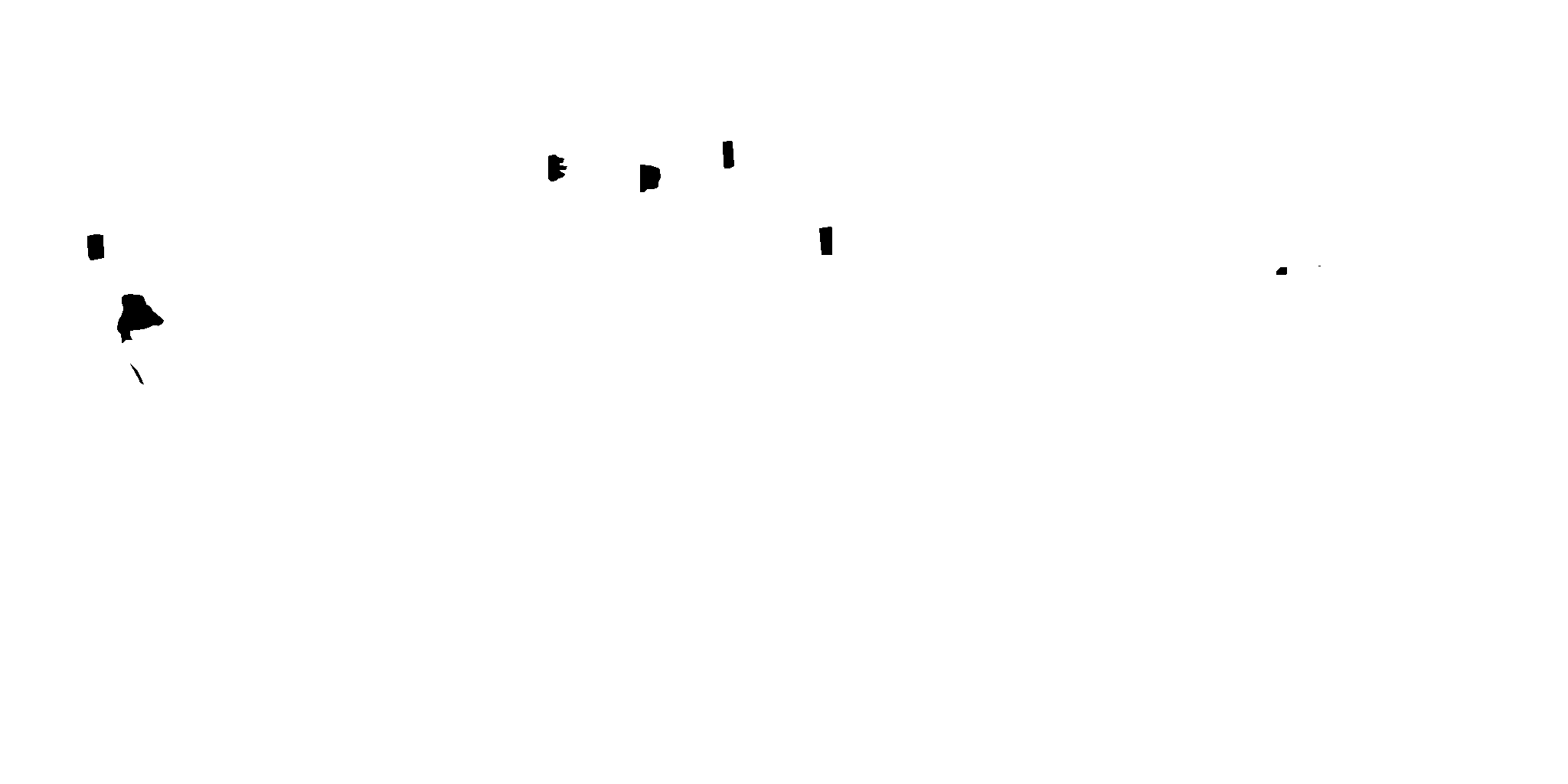}}}\\
\columnname{\textbf{Vistas}}\\
\subfloat{\includegraphics[width=\tempwidth,height=\tempheight]{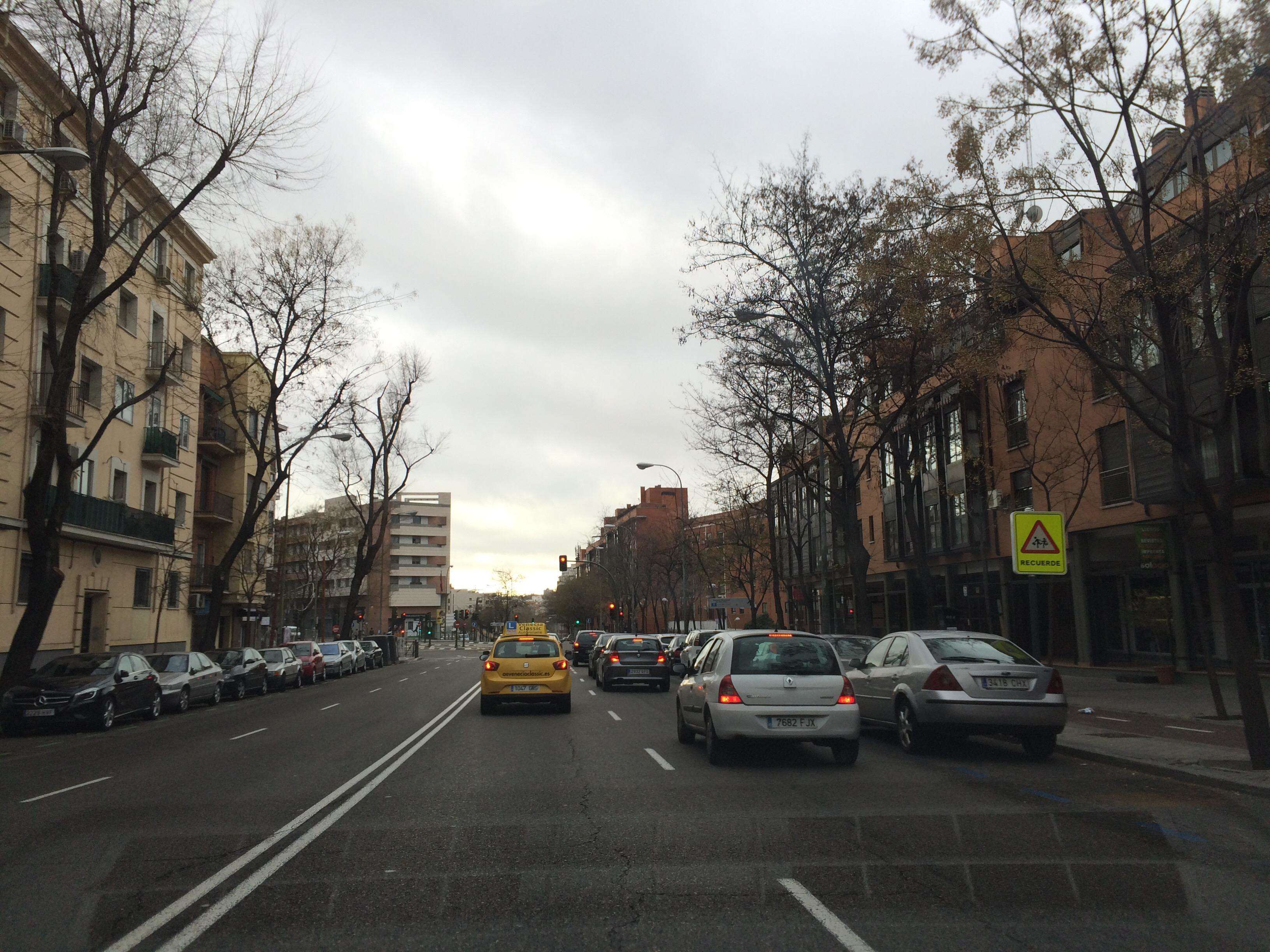}}\hfil
\subfloat{\includegraphics[width=\tempwidth,height=\tempheight]{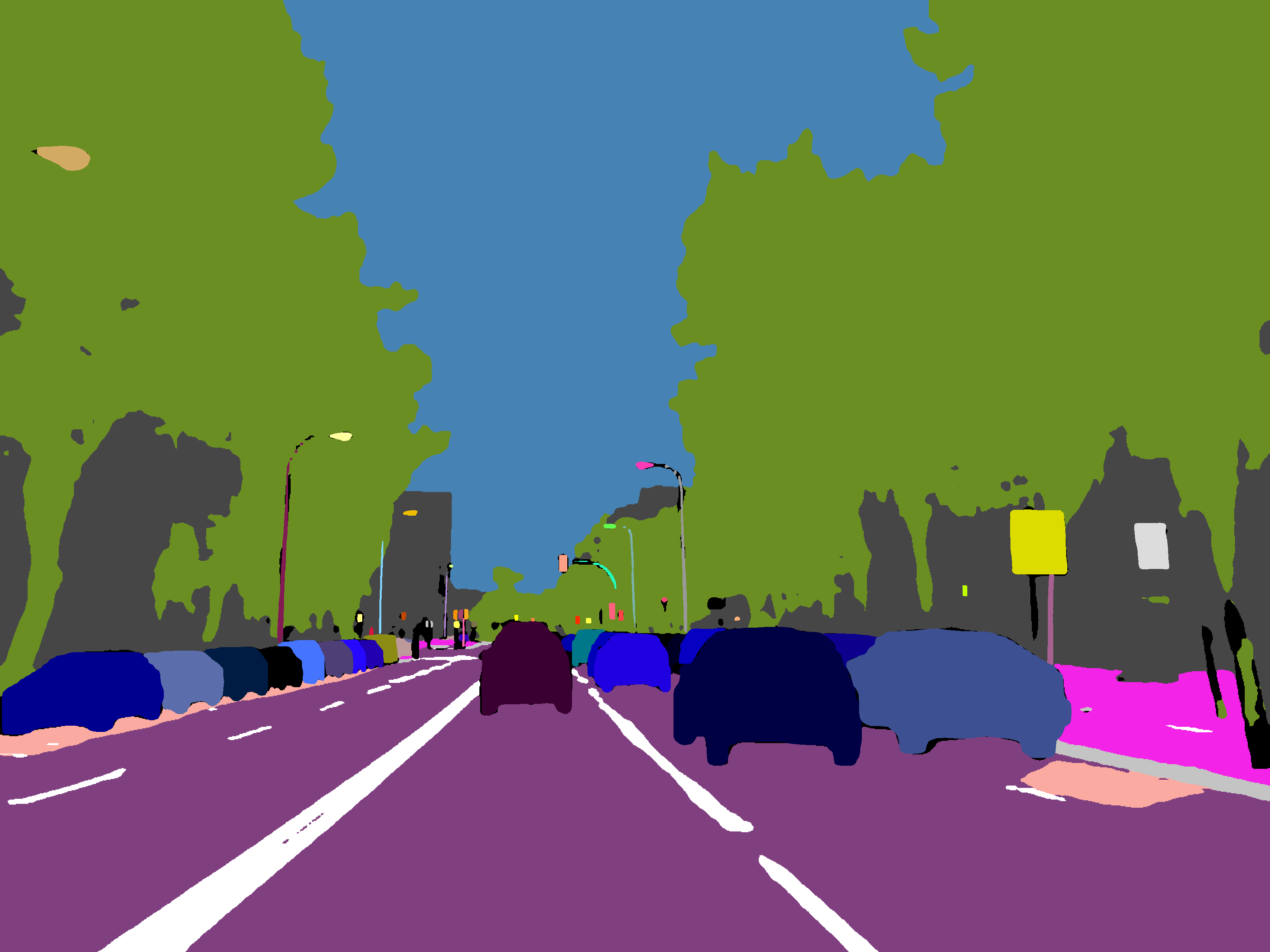}}\hfil
\subfloat{\frame{\includegraphics[width=\tempwidth,height=\tempheight]{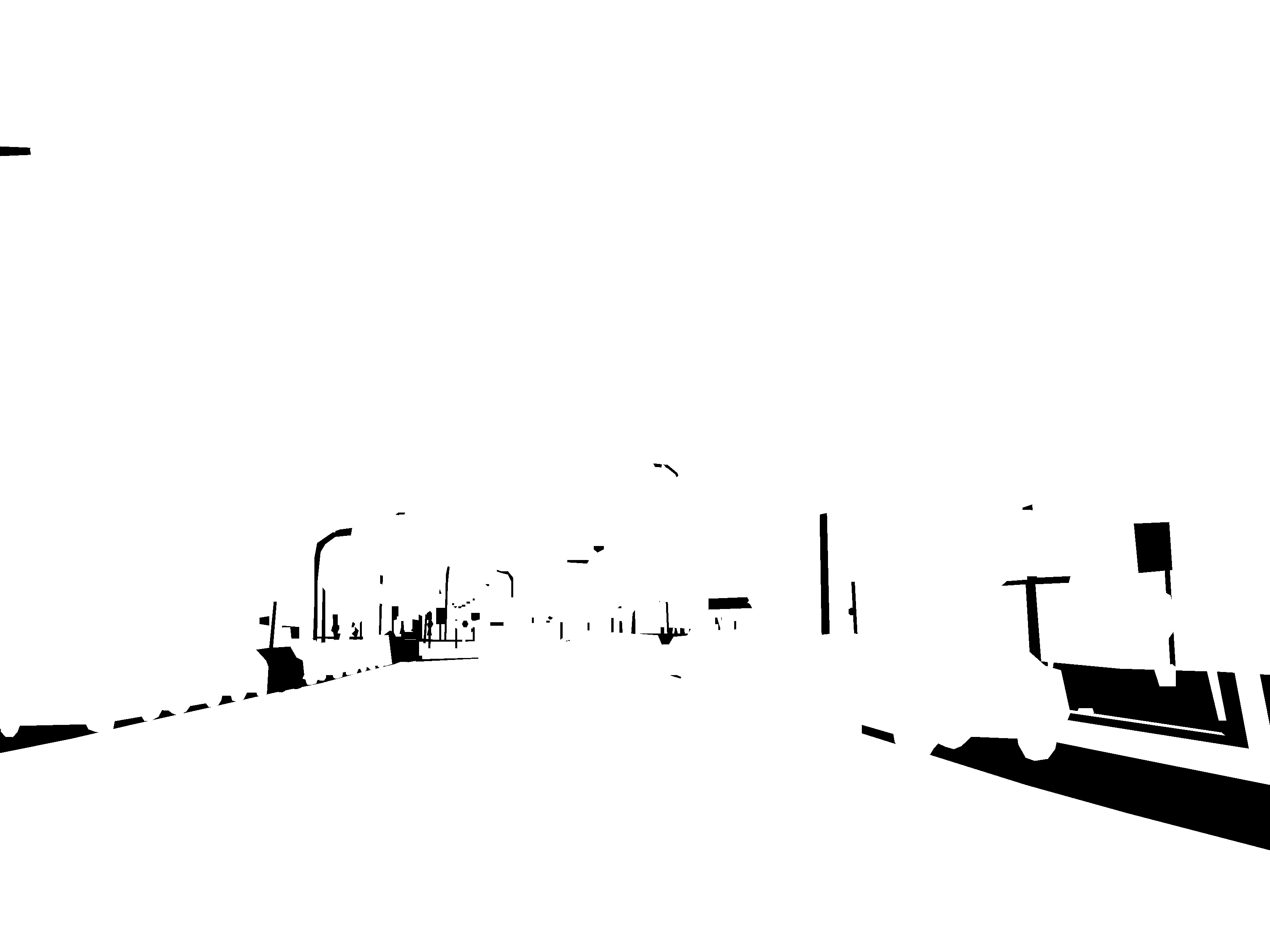}}}\\
\subfloat{\includegraphics[width=\tempwidth,height=\tempheight]{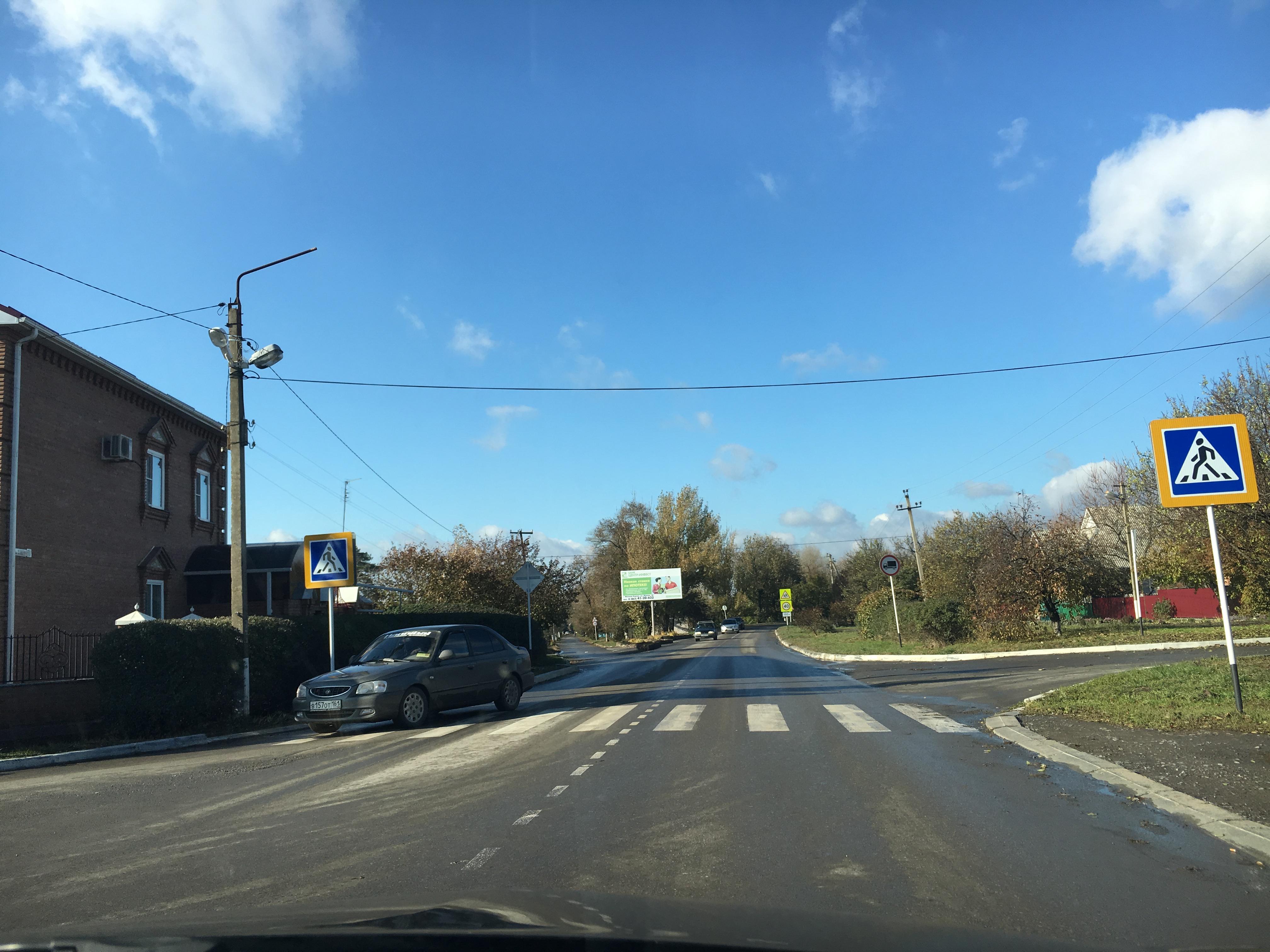}}\hfil
\subfloat{\includegraphics[width=\tempwidth,height=\tempheight]{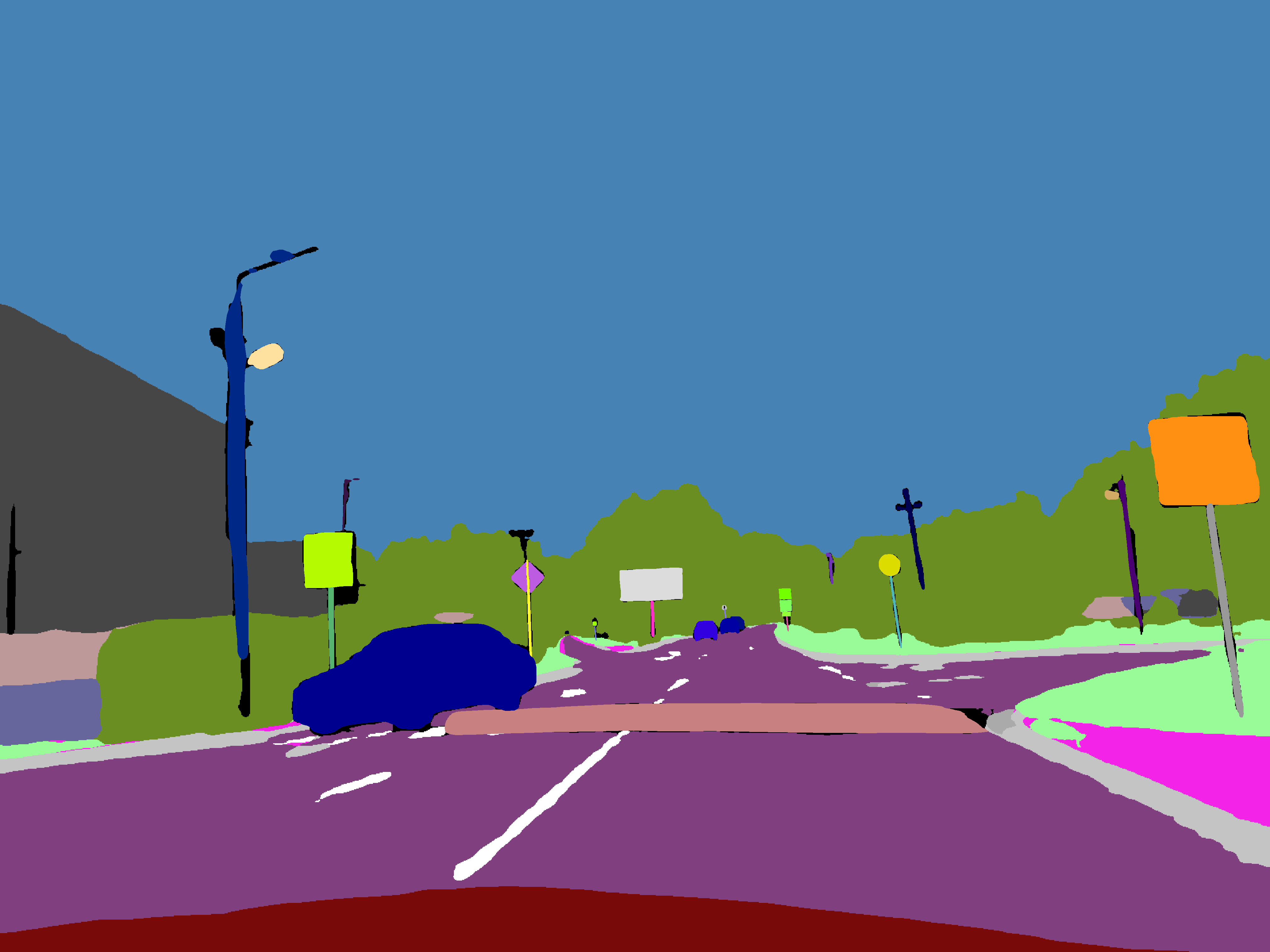}}\hfil
\subfloat{\frame{\includegraphics[width=\tempwidth,height=\tempheight]{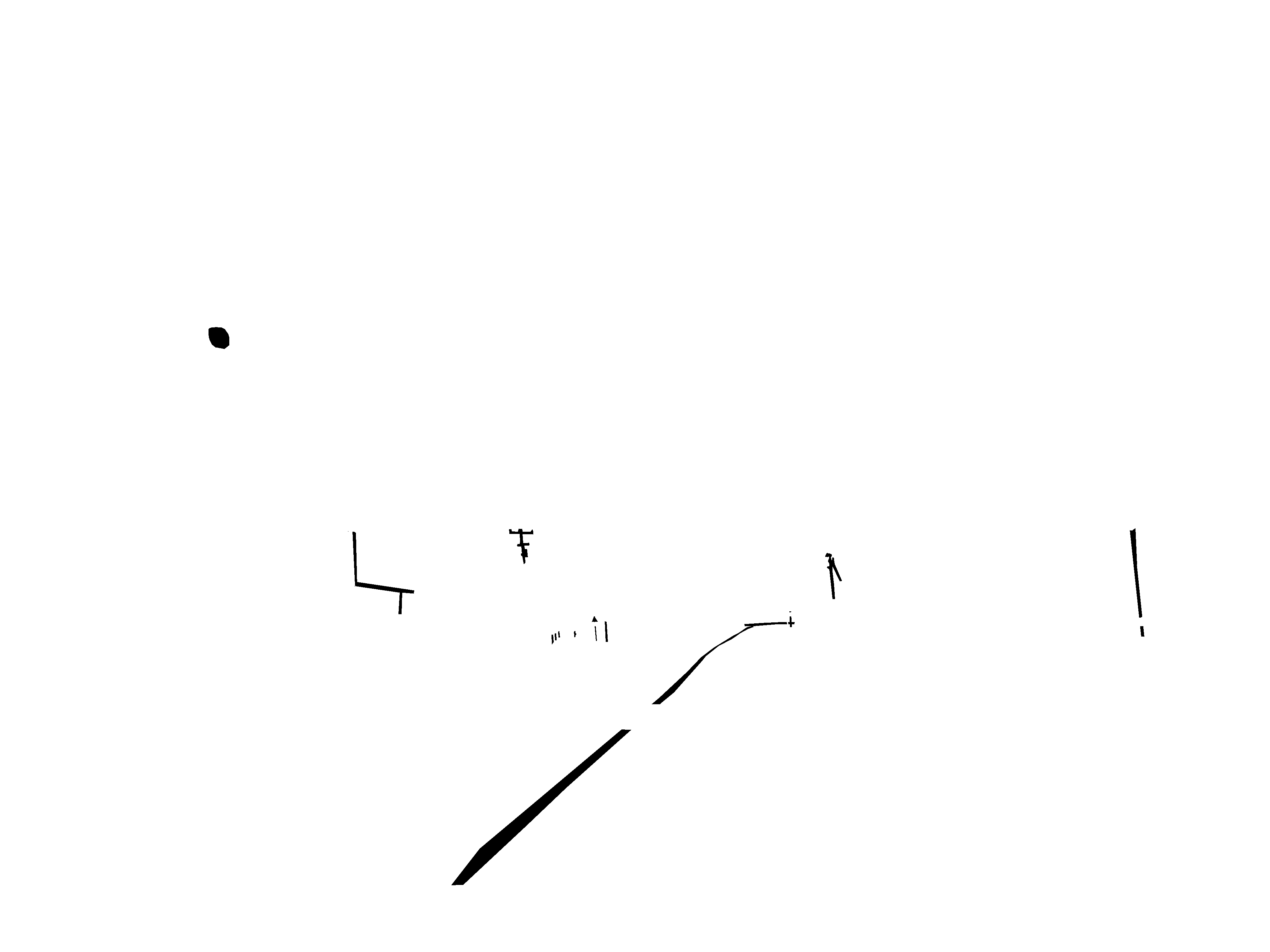}}}\\

\columnname{Raw Image}\hfil
\columnname{TASCNet Panoptic Segmentation}\hfil
\columnname{Mismatched Segments}\\
\caption{\textbf{Panoptic segmentation examples from Cityscapes and Mapillary Vistas.} In panoptic segmentation results, different instances are color-coded with different colors with small variation from the base color of their semantic class. In matched segments, segments belongs to true positives are marked as white, while false positive and false negative segments are marked as black. }
\label{fig:result}
\end{figure*}



\section{Conclusion}
In this work, we proposed an end-to-end network, TASCNet, to jointly predict stuff and things. We demonstrate that a novel cross-task constraint can boost performance on instance, semantic, and panoptic segmentation performance. Our proposed approach is competitive with state-of-the-art models on both panoptic segmentation task and single modal segmentation tasks on benchmark datasets, while using far fewer parameters. 


{
\bibliographystyle{ieee}
\bibliography{reference/adrien,reference/triplenet}
}
\clearpage
\balance
\section*{Appendix: Detailed Experimental Results\label{sec:appendix}}
\begin{figure*}[htb!]
\setlength{\tempwidth}{.32\linewidth}
\setlength{\tempheight}{0.16\linewidth}
\centering
\subfloat{\includegraphics[width=\tempwidth,height=\tempheight]{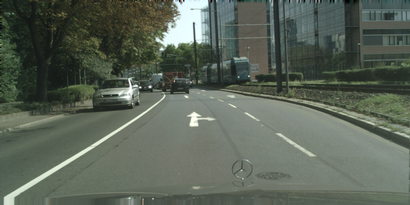}}\hfil
\subfloat{\includegraphics[width=\tempwidth,height=\tempheight]{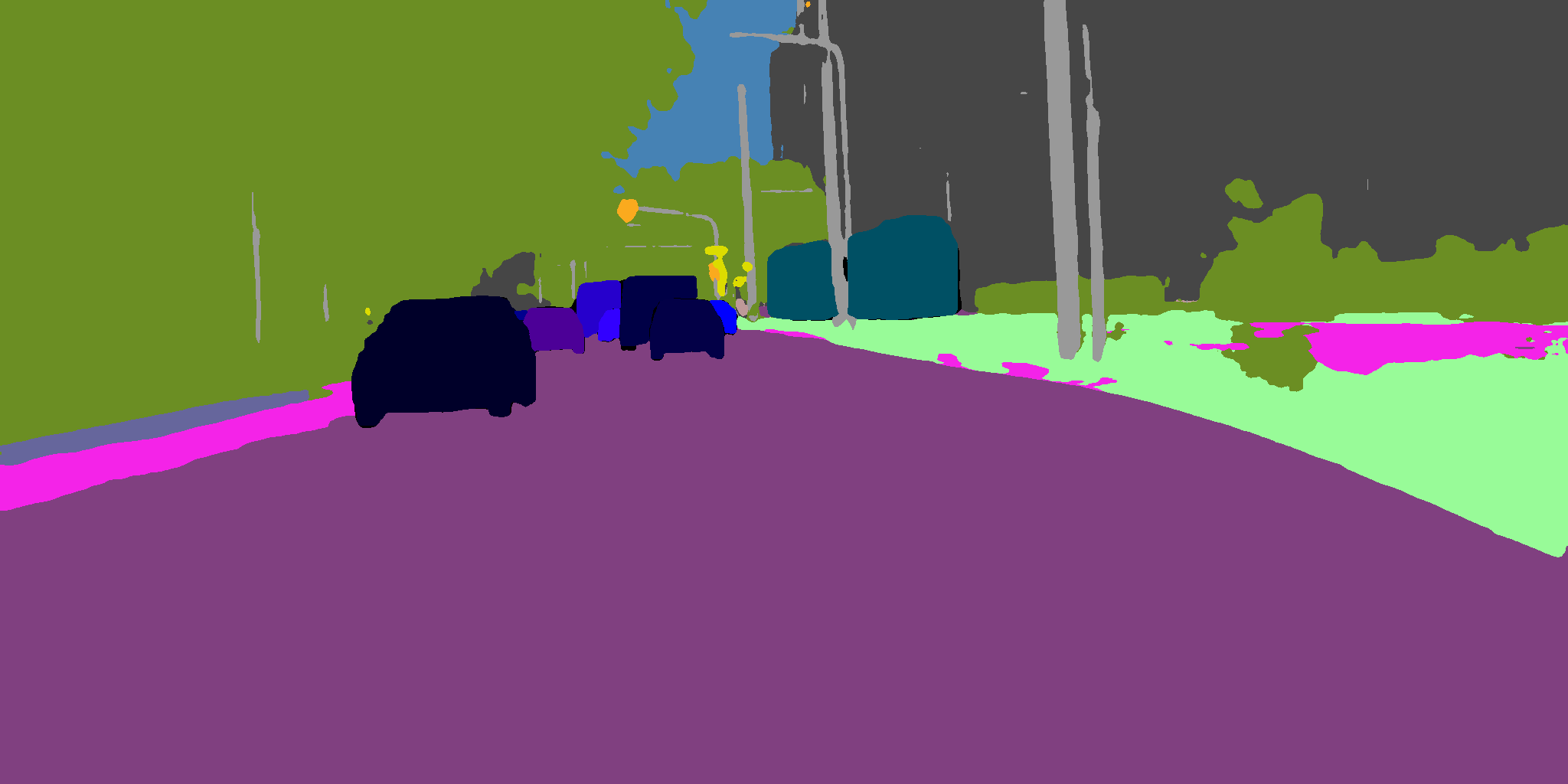}}\hfil
\subfloat{\frame{\includegraphics[width=\tempwidth,height=\tempheight]{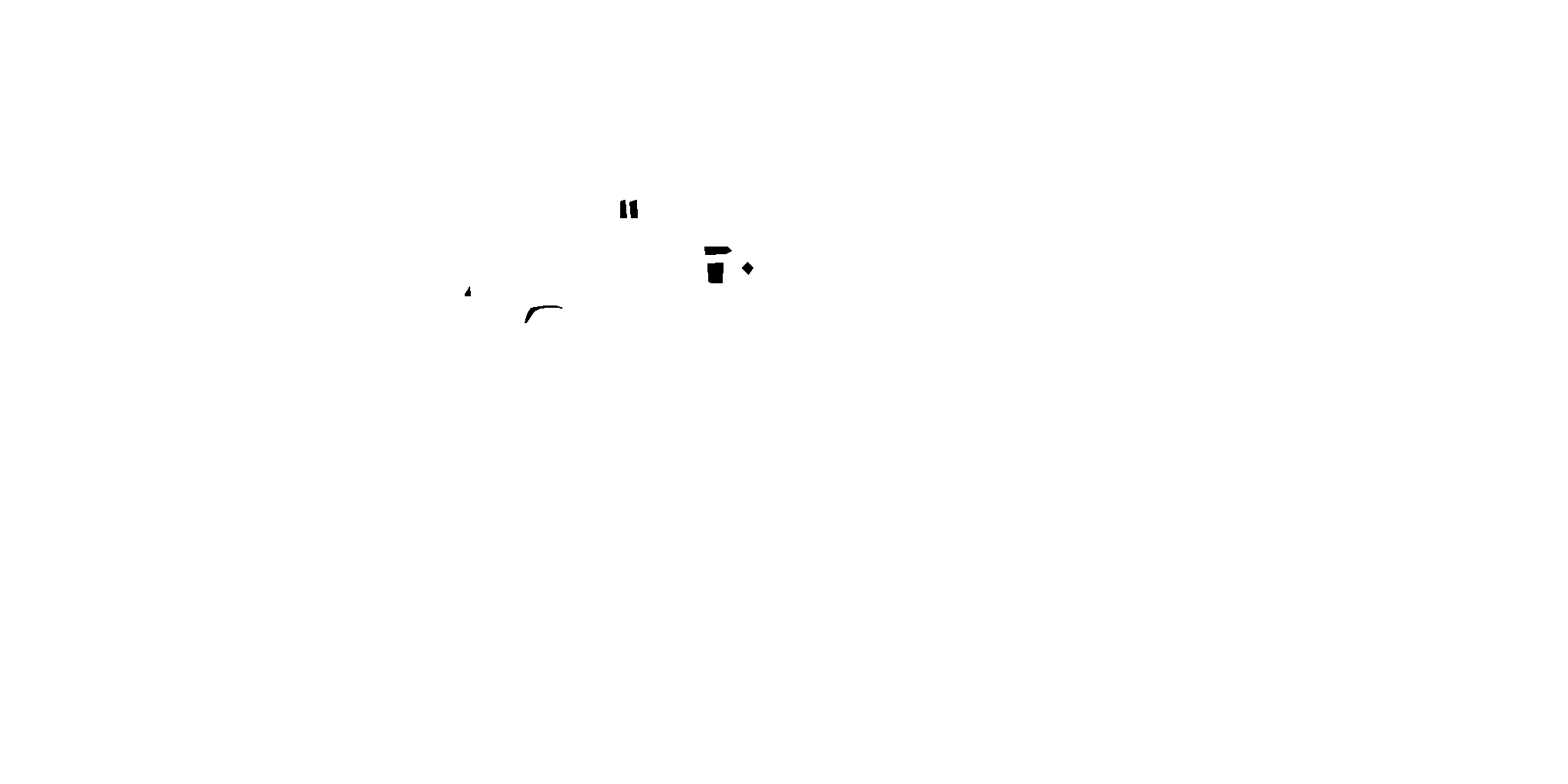}}}\\
\subfloat{\includegraphics[width=\tempwidth,height=\tempheight]{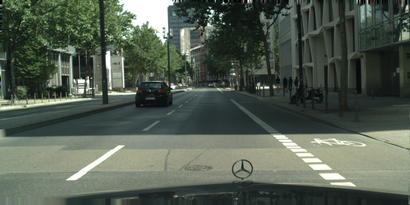}}\hfil
\subfloat{\includegraphics[width=\tempwidth,height=\tempheight]{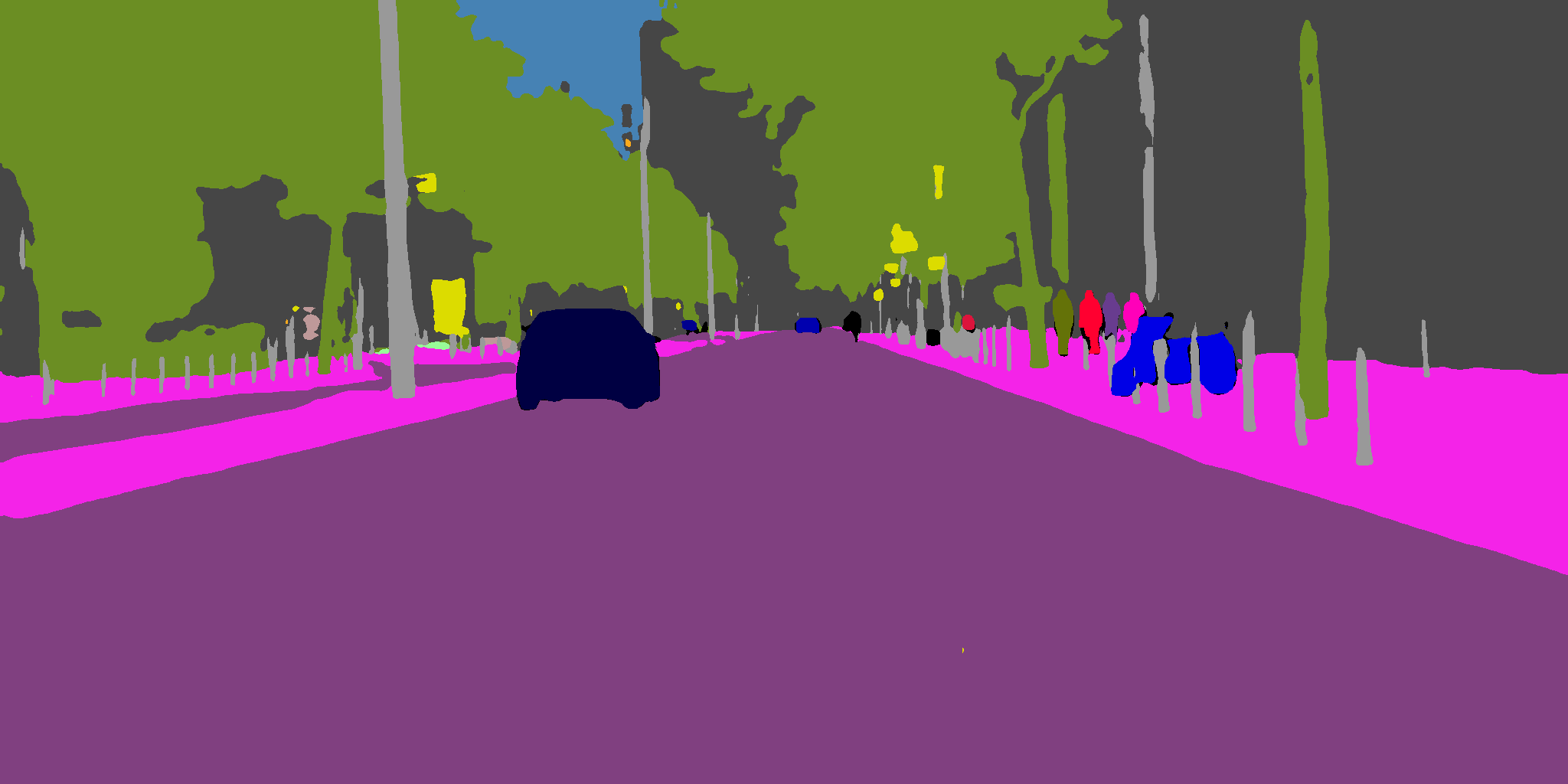}}\hfil
\subfloat{\frame{\includegraphics[width=\tempwidth,height=\tempheight]{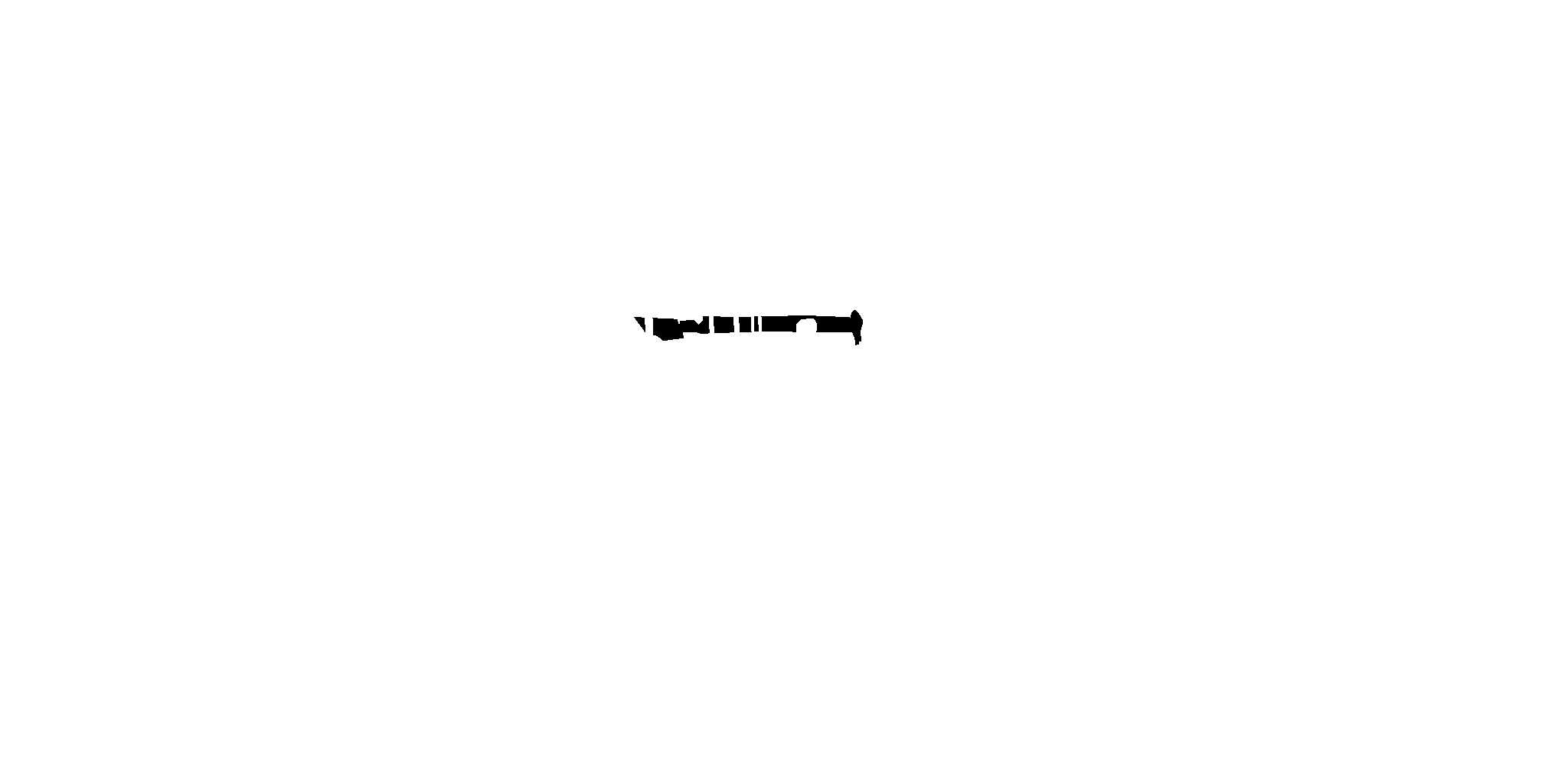}}}\\
\subfloat{\includegraphics[width=\tempwidth,height=\tempheight]{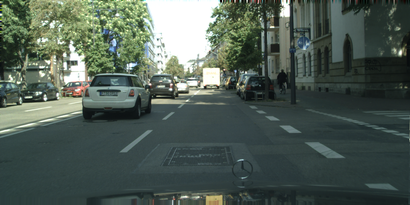}}\hfil
\subfloat{\includegraphics[width=\tempwidth,height=\tempheight]{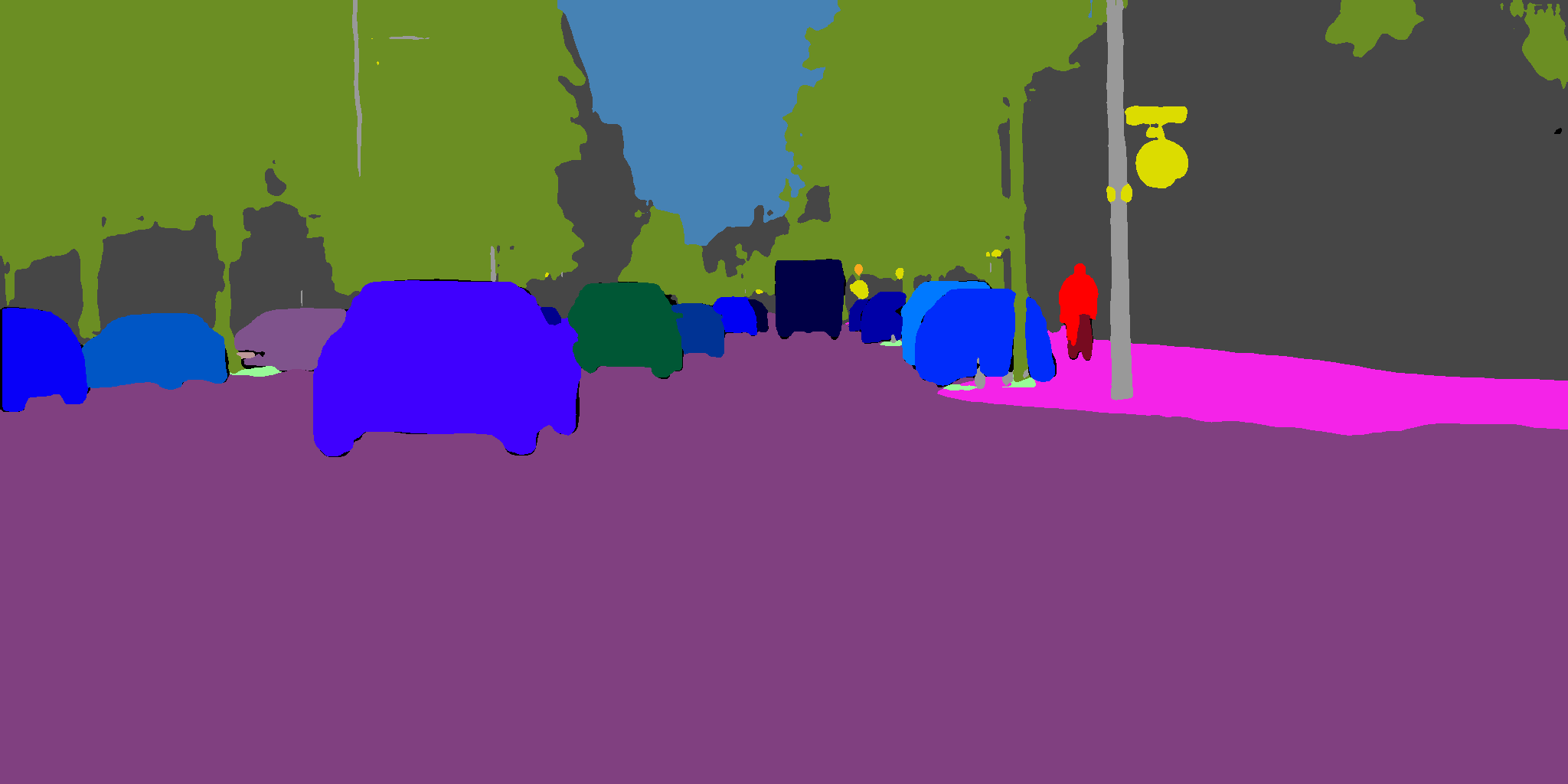}}\hfil
\subfloat{\frame{\includegraphics[width=\tempwidth,height=\tempheight]{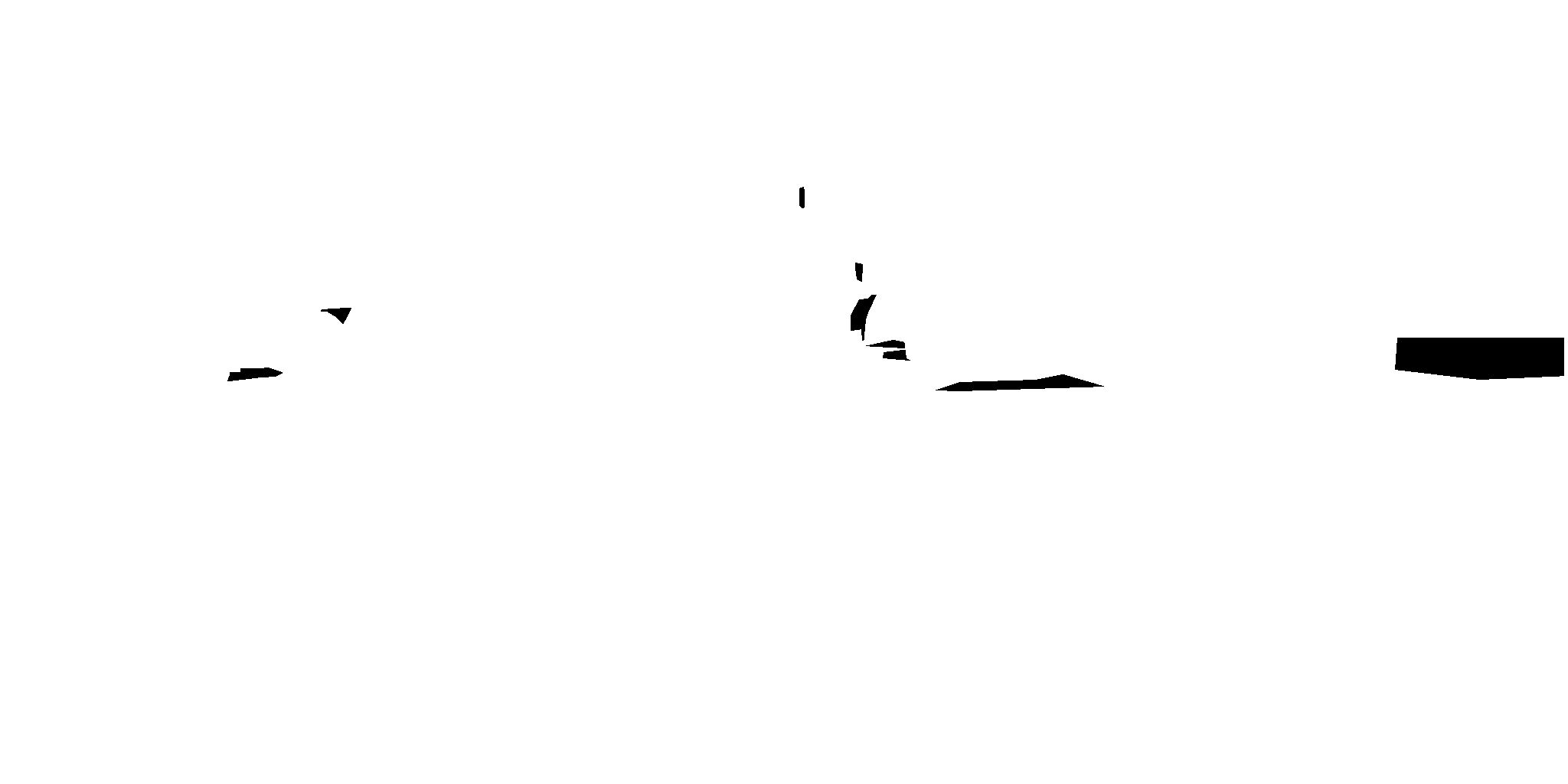}}}\\
\subfloat{\includegraphics[width=\tempwidth,height=\tempheight]{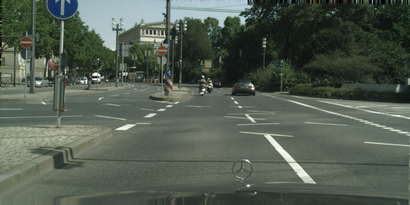}}\hfil
\subfloat{\includegraphics[width=\tempwidth,height=\tempheight]{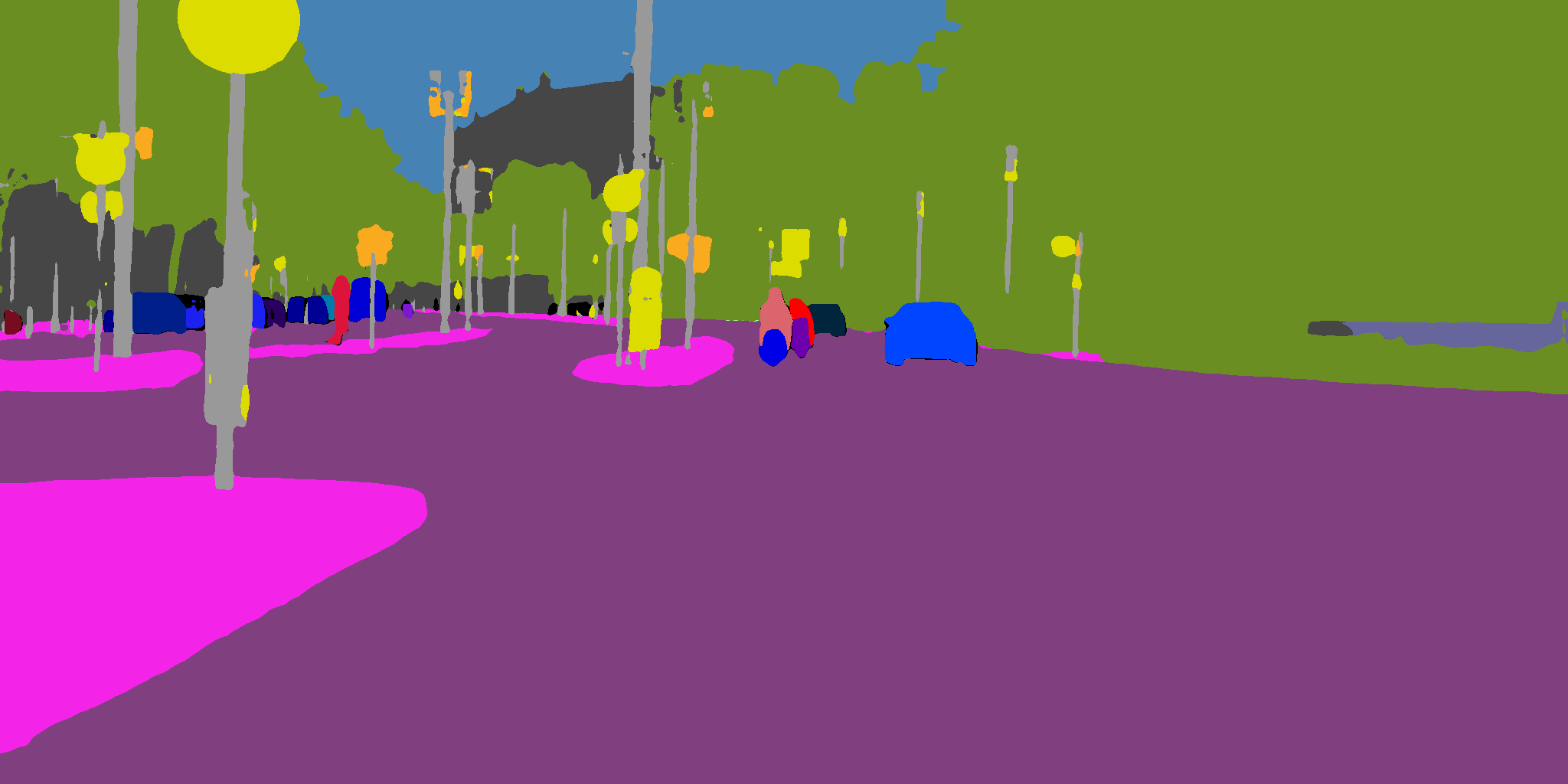}}\hfil
\subfloat{\frame{\includegraphics[width=\tempwidth,height=\tempheight]{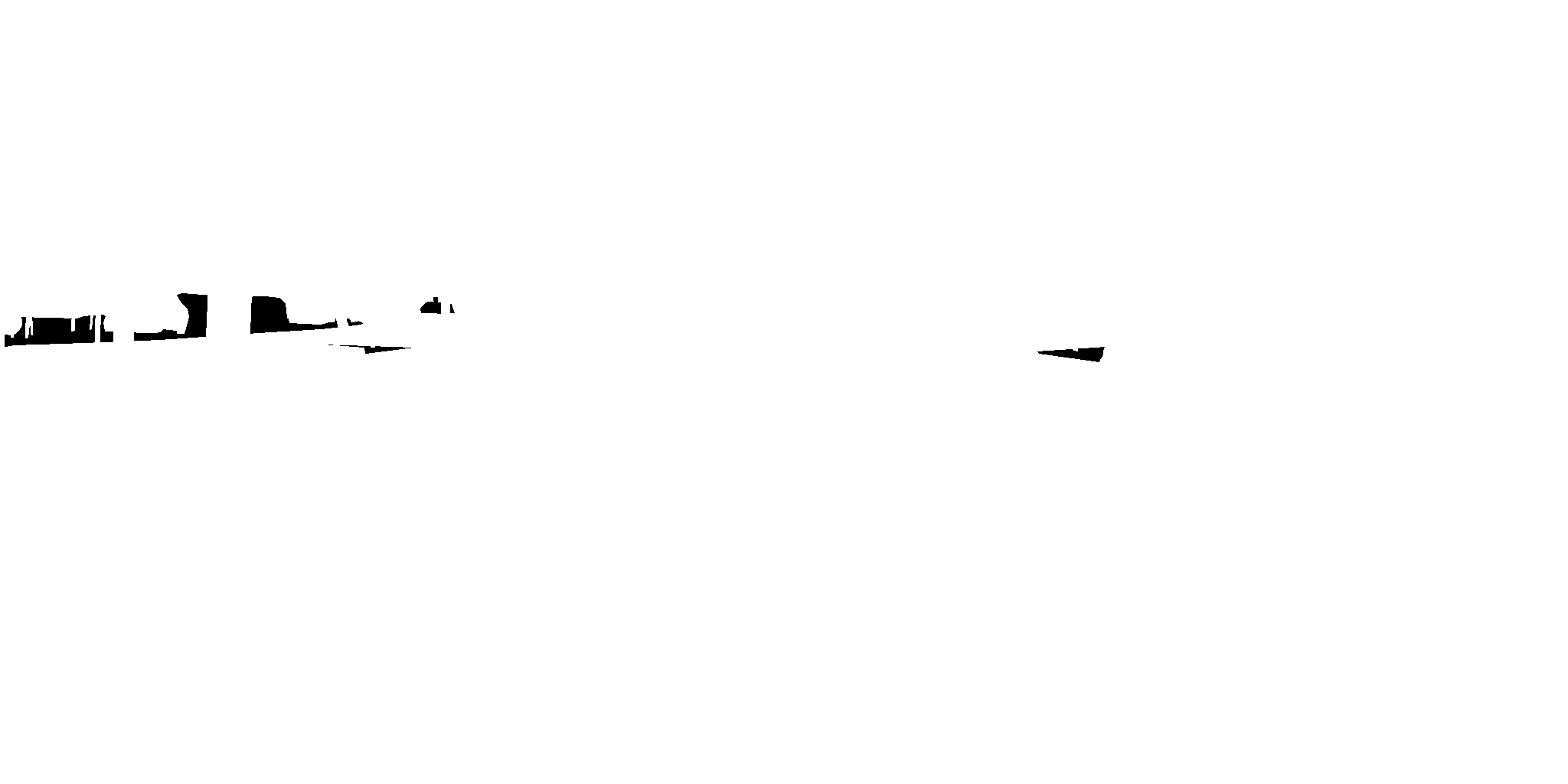}}}\\
\columnname{Raw Image}\hfil
\columnname{TASCNet Panoptic Segmentation}\hfil
\columnname{Mismatched Segments}\\
\caption{\textbf{More Panoptic Segmentation Examples from Cityscapes.} In panoptic segmentation results, different instances are color-coded with different colors with small variations from the base color of their semantic class. In mismatched segments, segments belongs to true positives are marked as white, while false positive and false negative segments are marked as black.}
\label{fig:cityscapes-results}
\end{figure*}

\begin{figure*}[htb!]
\setlength{\tempwidth}{.32\linewidth}
\setlength{\tempheight}{0.16\linewidth}
\centering
\subfloat{\includegraphics[width=\tempwidth,height=\tempheight]{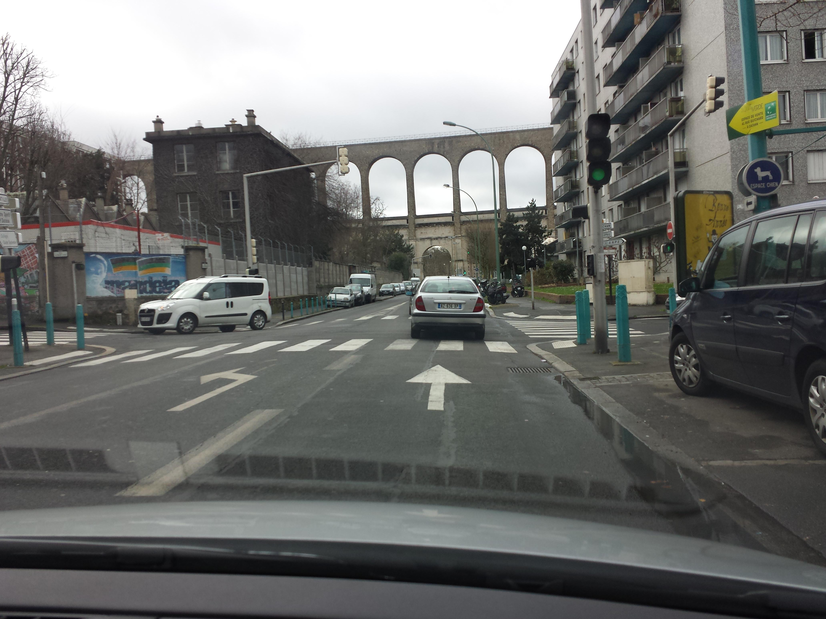}}\hfil
\subfloat{\includegraphics[width=\tempwidth,height=\tempheight]{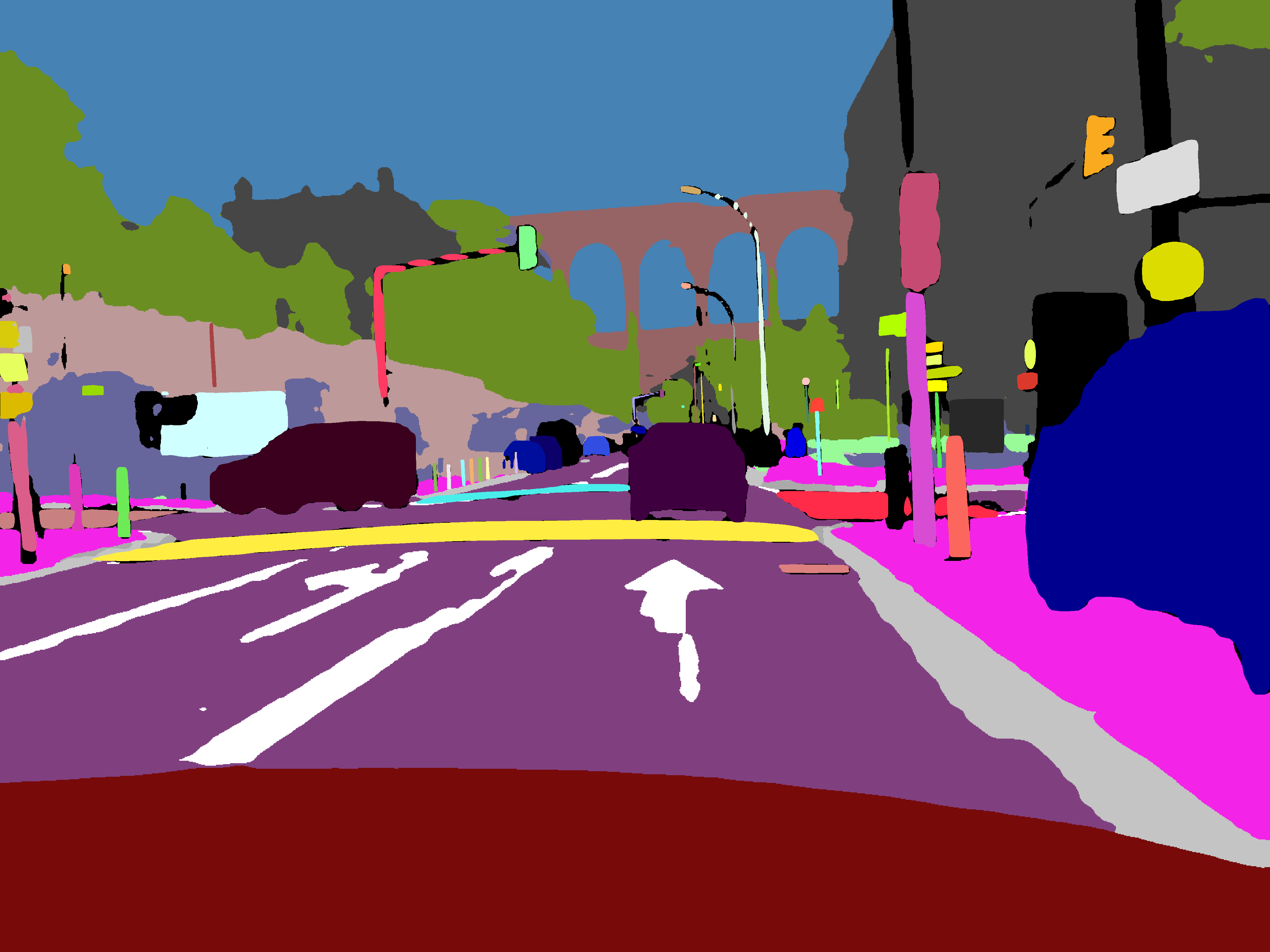}}\hfil
\subfloat{\frame{\includegraphics[width=\tempwidth,height=\tempheight]{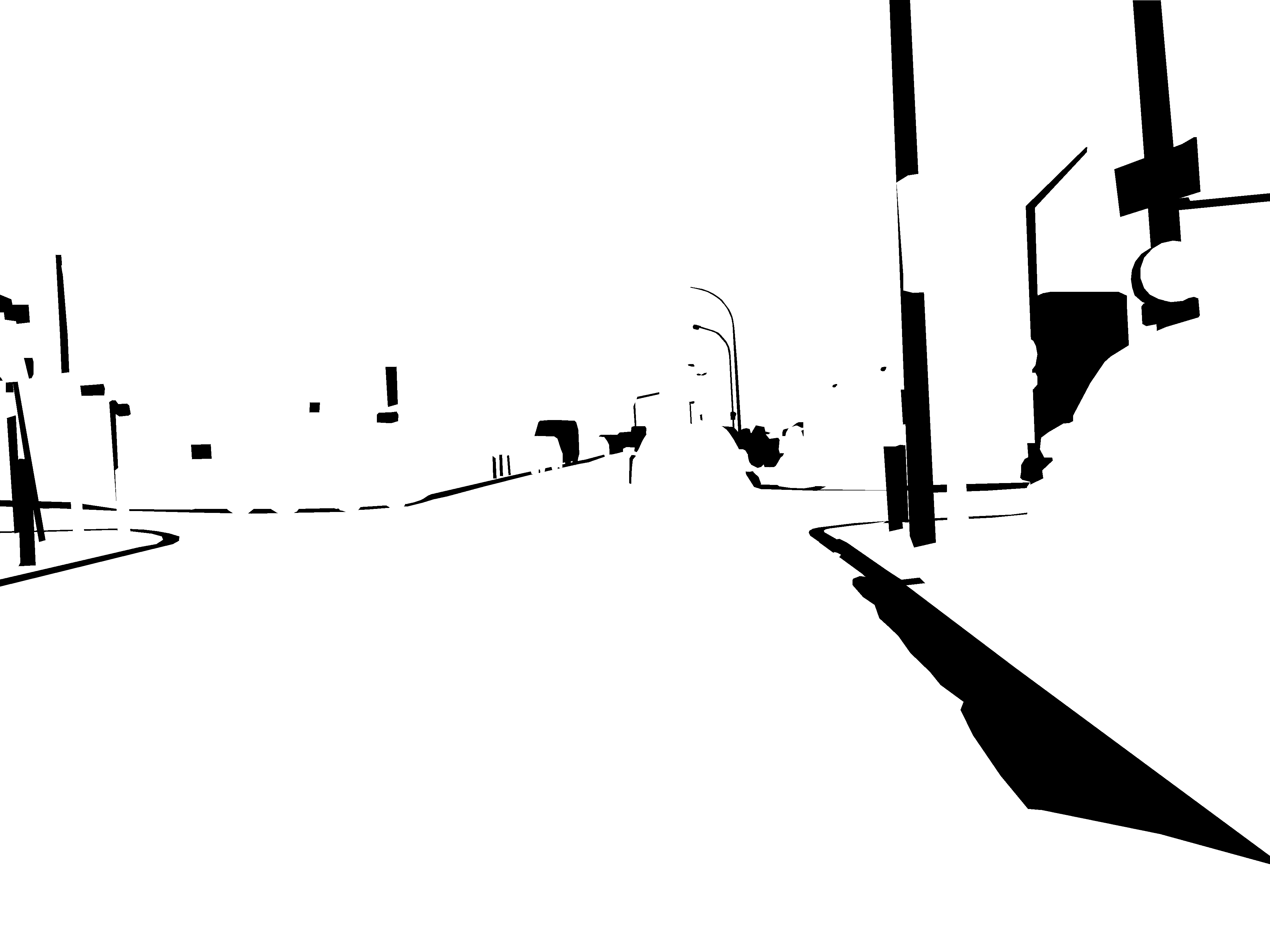}}}\\
\subfloat{\includegraphics[width=\tempwidth,height=\tempheight]{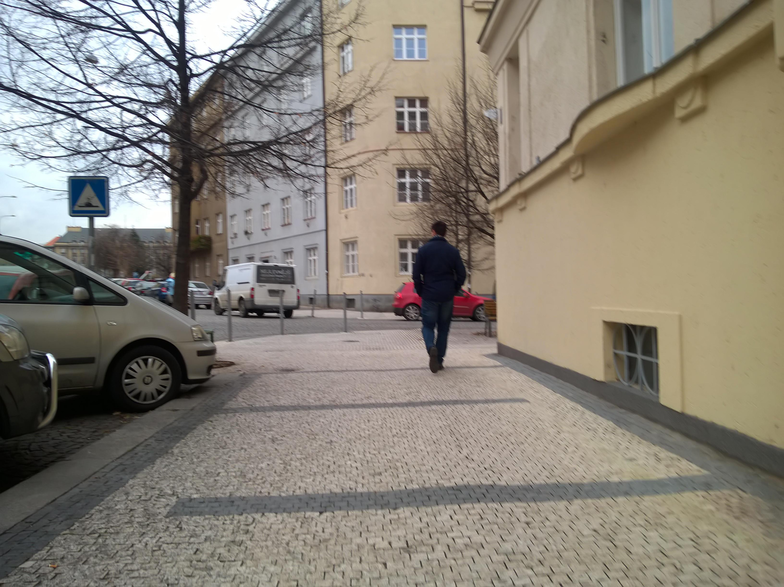}}\hfil
\subfloat{\includegraphics[width=\tempwidth,height=\tempheight]{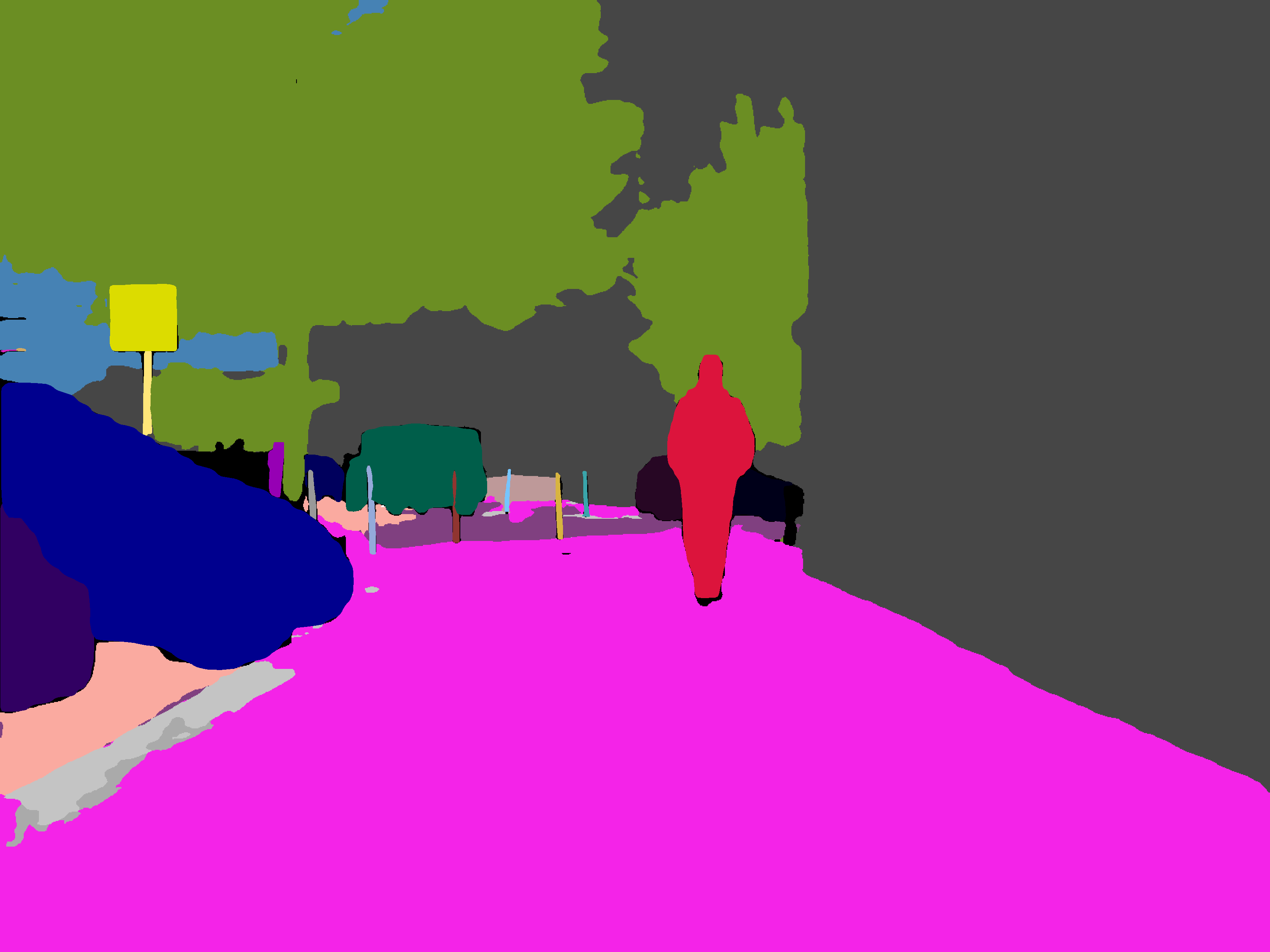}}\hfil
\subfloat{\frame{\includegraphics[width=\tempwidth,height=\tempheight]{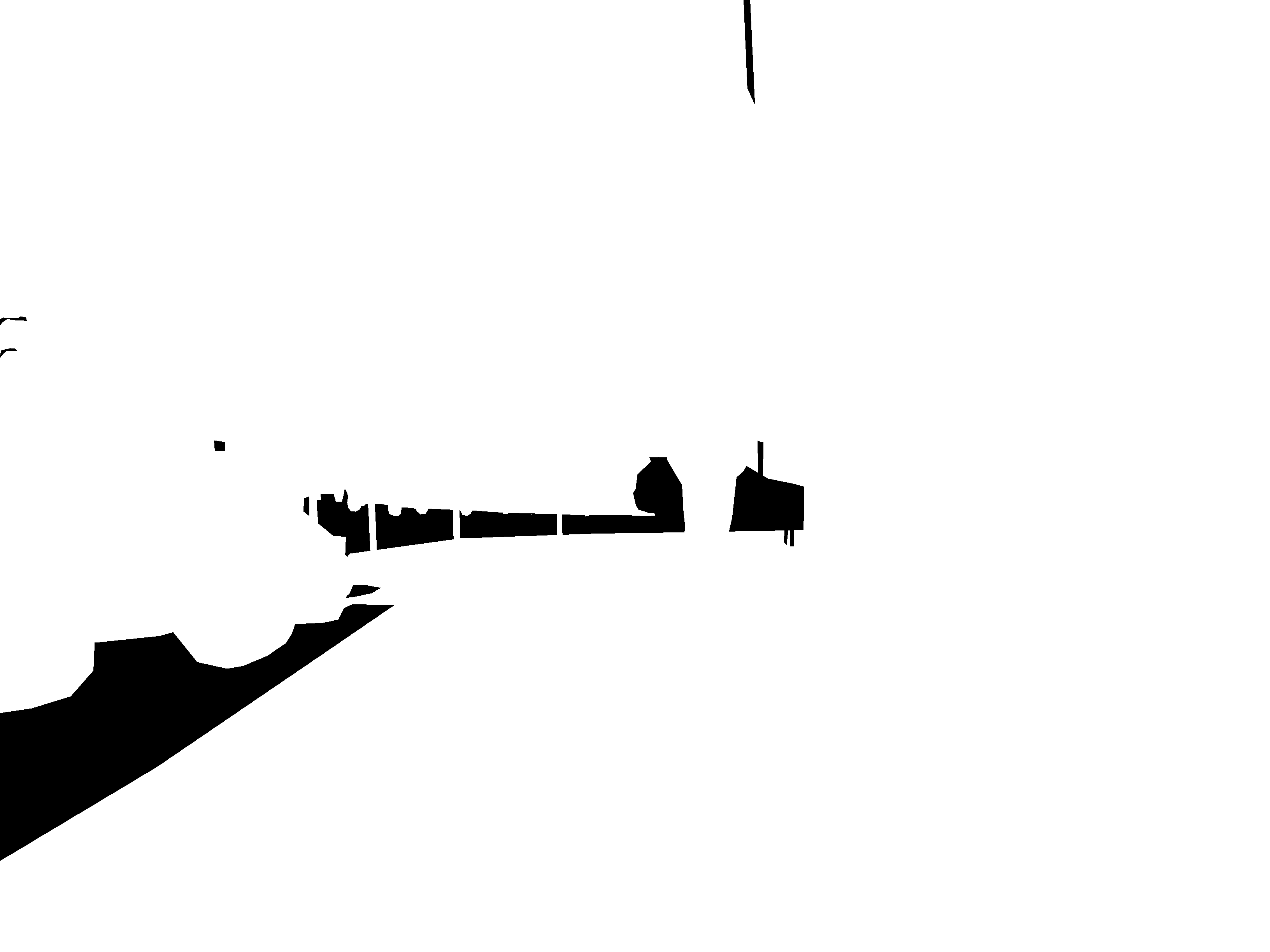}}}\\
\subfloat{\includegraphics[width=\tempwidth,height=\tempheight]{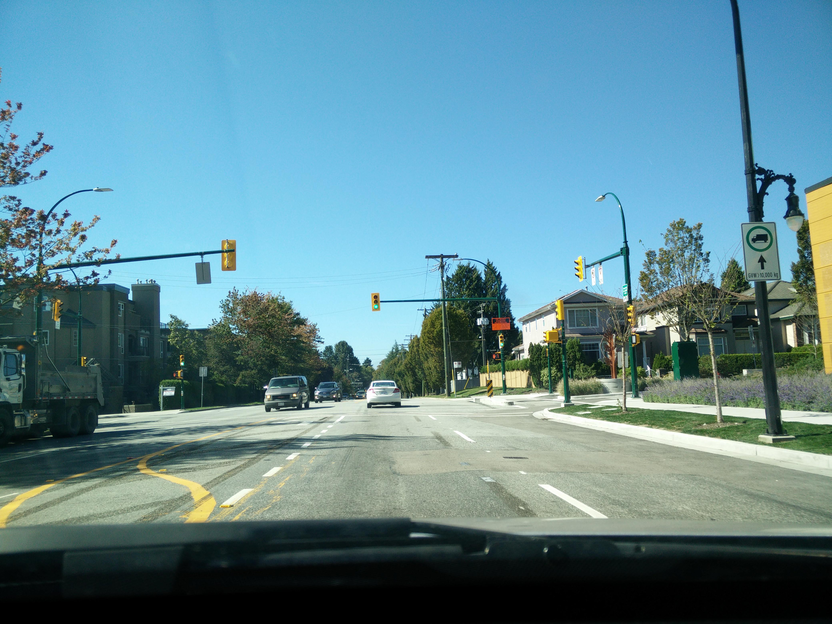}}\hfil
\subfloat{\includegraphics[width=\tempwidth,height=\tempheight]{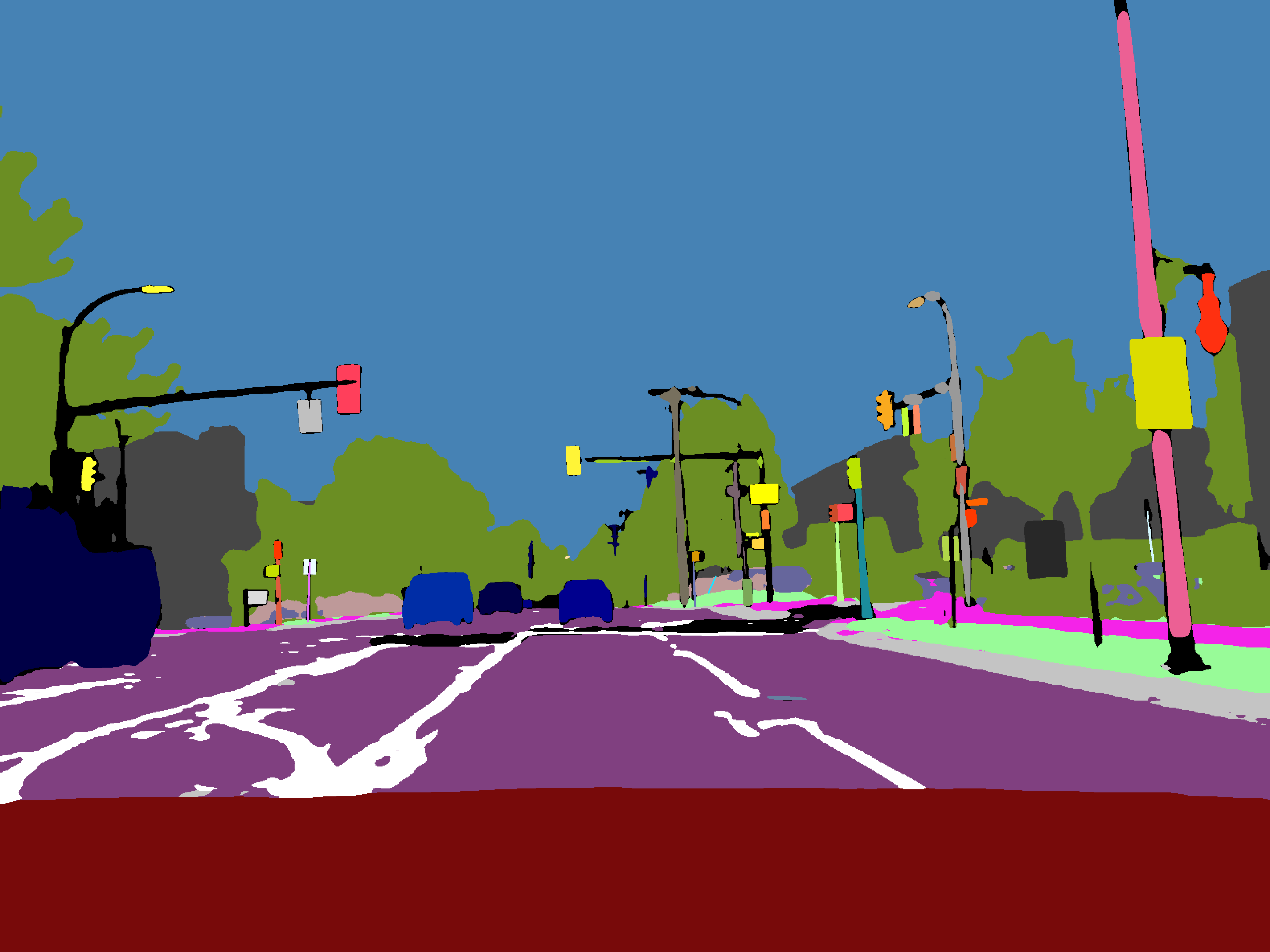}}\hfil
\subfloat{\frame{\includegraphics[width=\tempwidth,height=\tempheight]{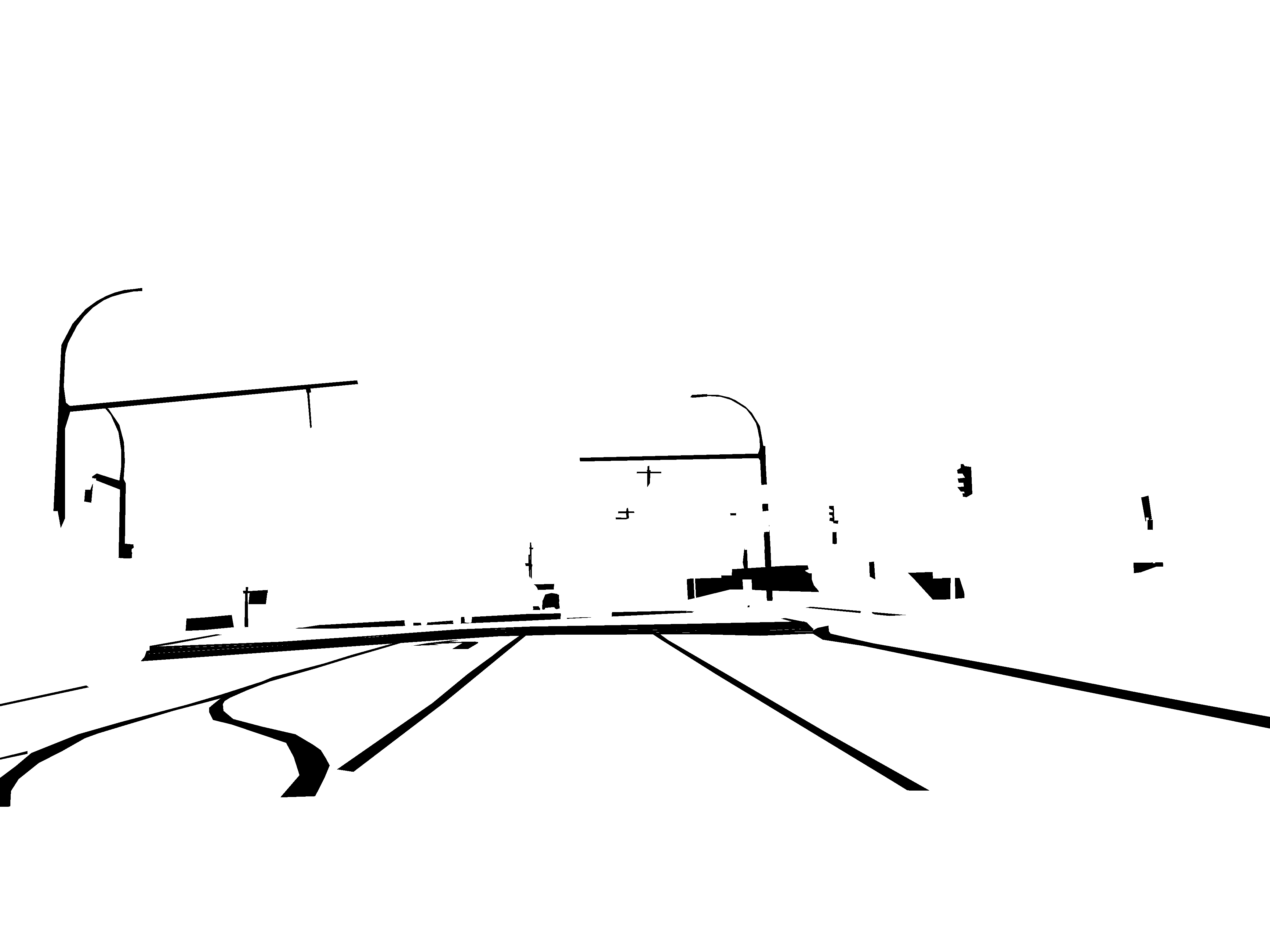}}}\\
\subfloat{\includegraphics[width=\tempwidth,height=\tempheight]{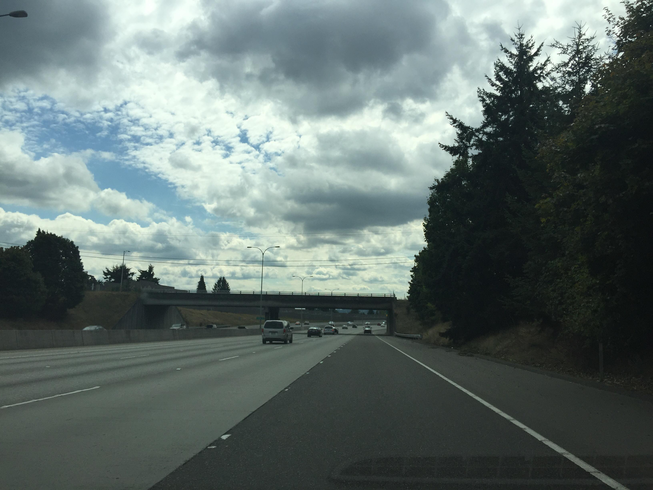}}\hfil
\subfloat{\includegraphics[width=\tempwidth,height=\tempheight]{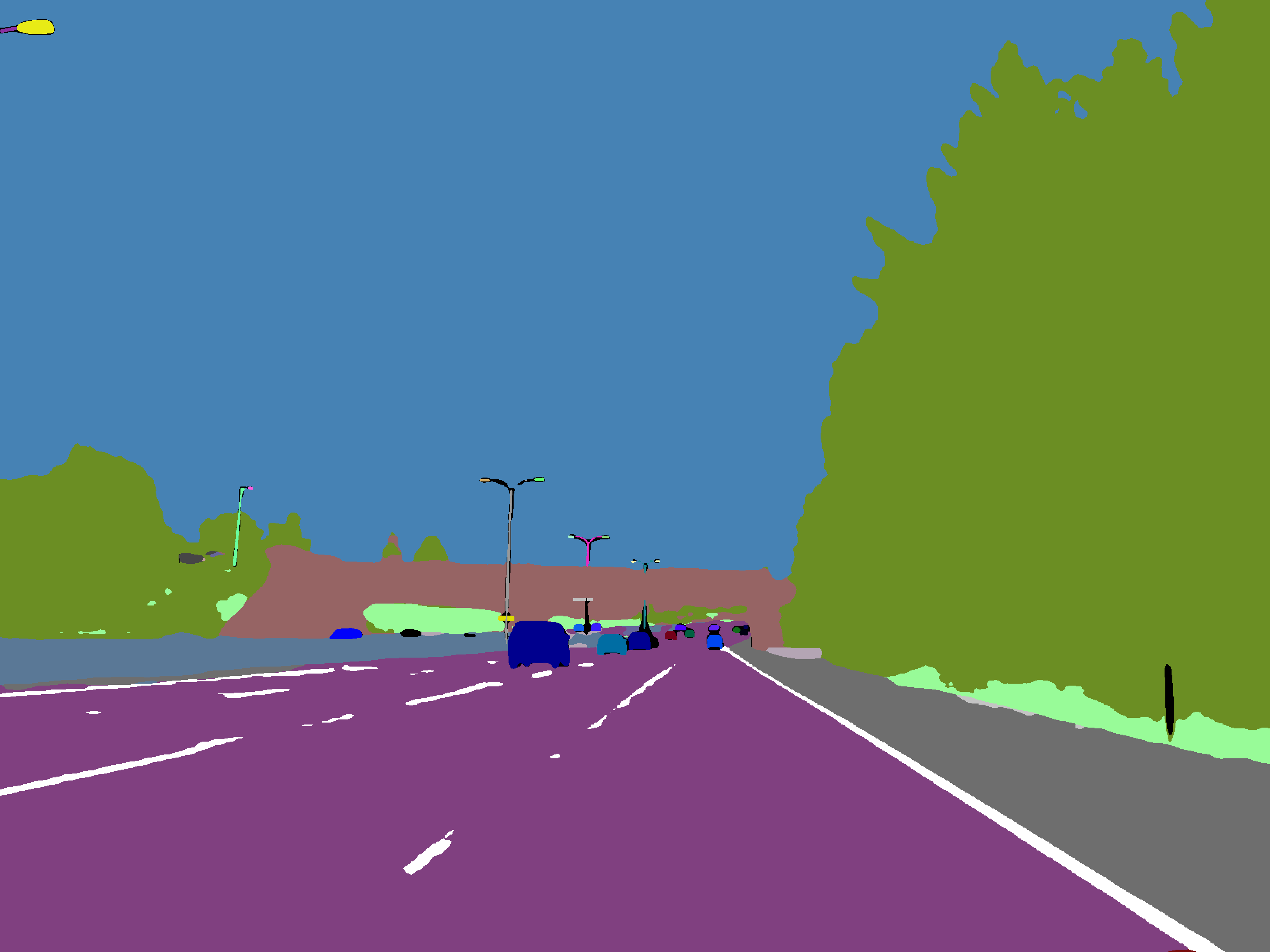}}\hfil
\subfloat{\frame{\includegraphics[width=\tempwidth,height=\tempheight]{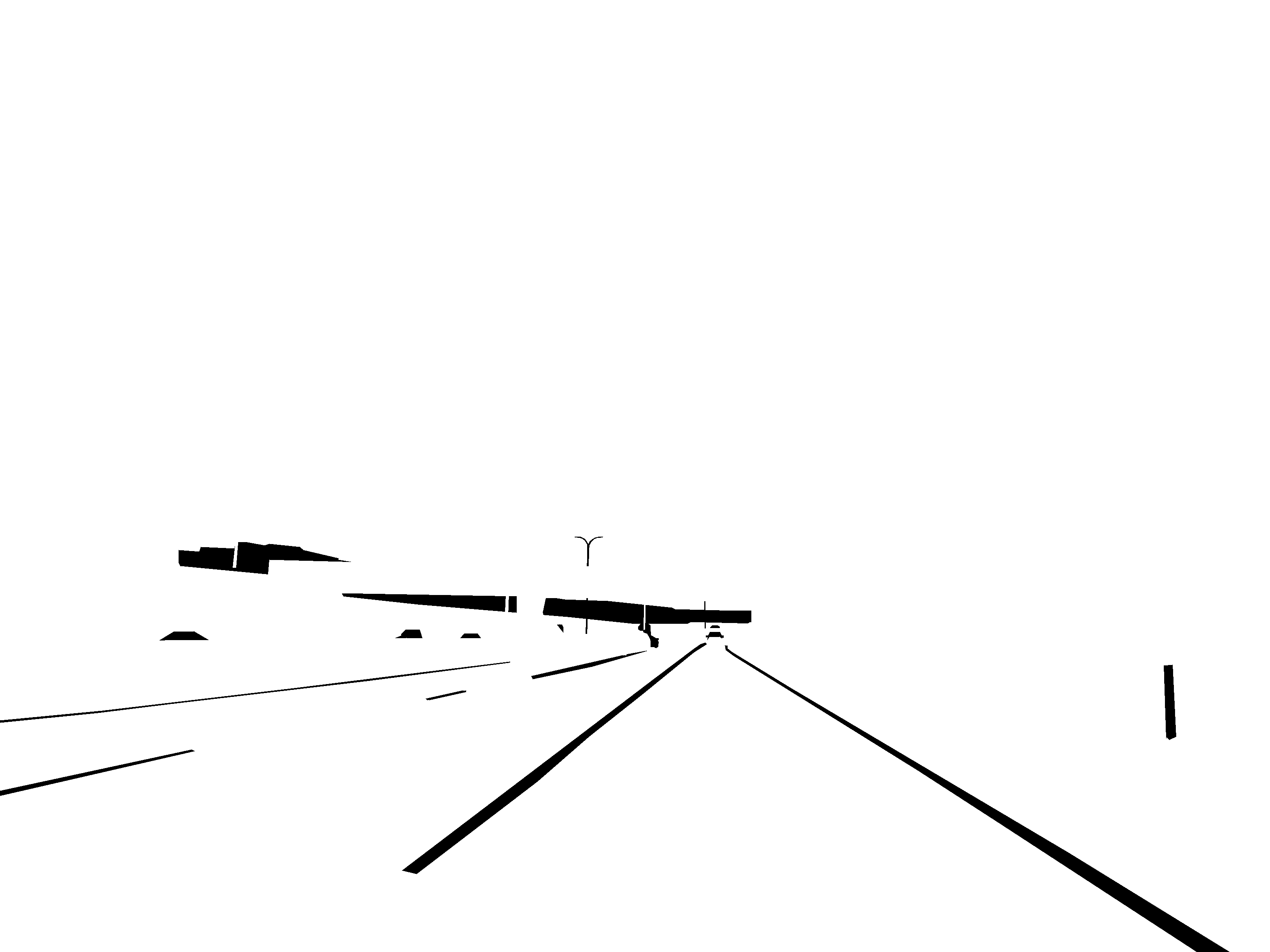}}}\\
\subfloat{\includegraphics[width=\tempwidth,height=\tempheight]{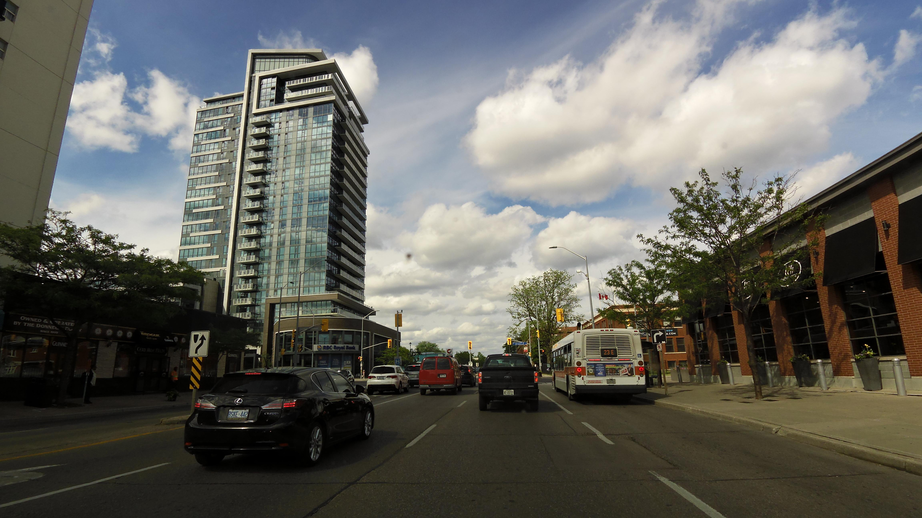}}\hfil
\subfloat{\includegraphics[width=\tempwidth,height=\tempheight]{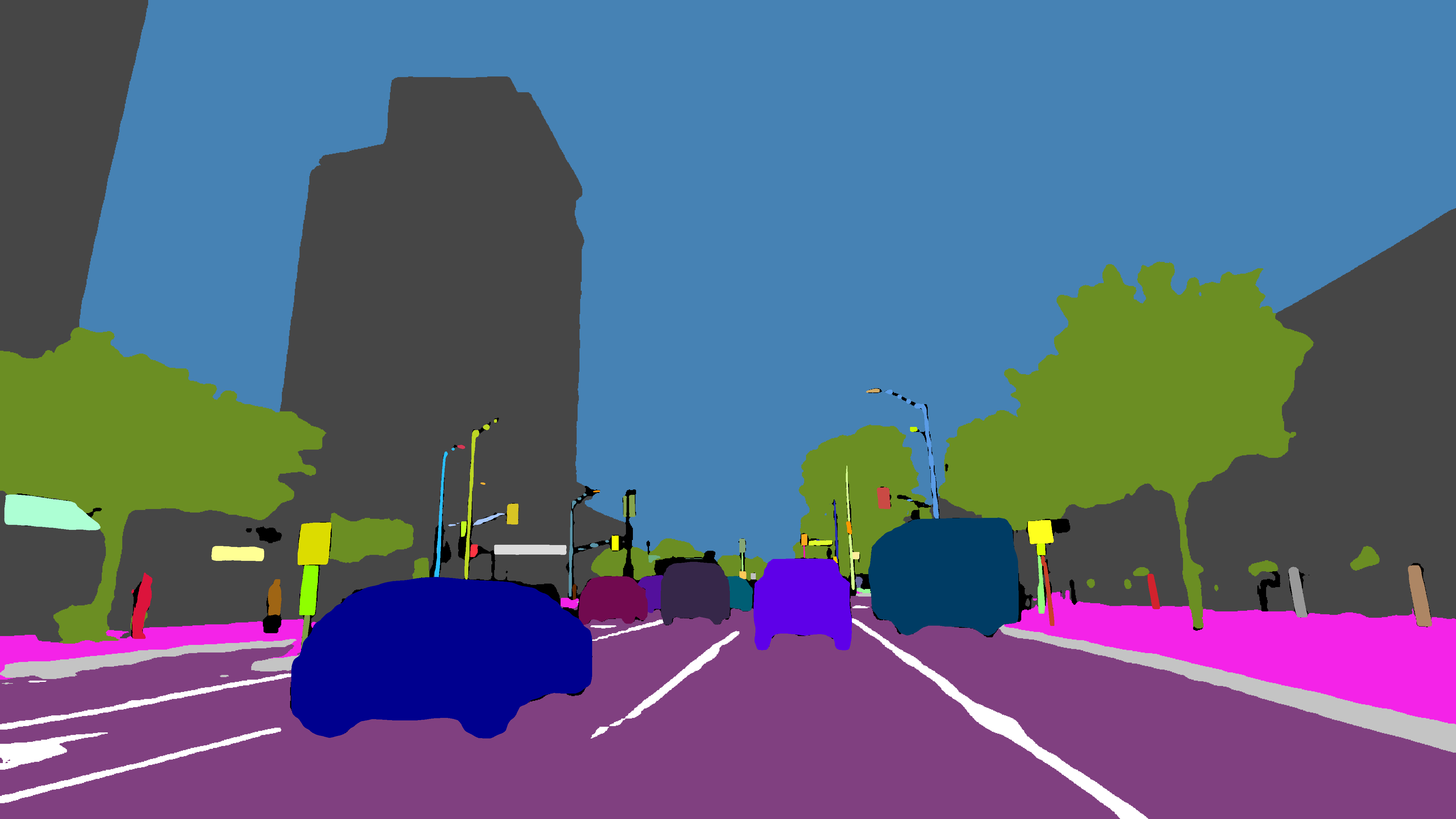}}\hfil
\subfloat{\frame{\includegraphics[width=\tempwidth,height=\tempheight]{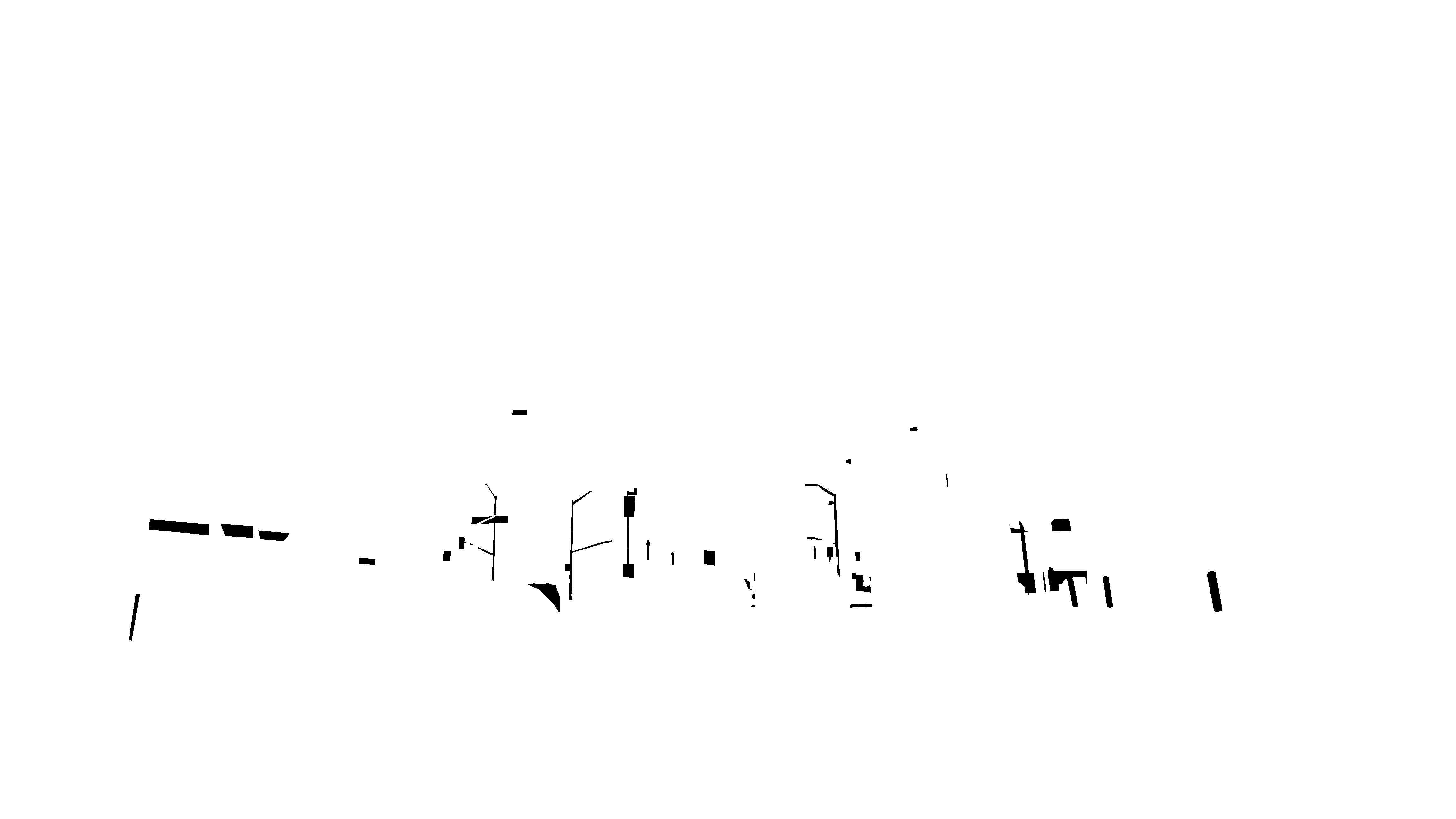}}}\\

\columnname{Raw Image}\hfil
\columnname{TASCNet Panoptic Segmentation}\hfil
\columnname{Mismatched Segments}\\
\caption{\textbf{More Panoptic Segmentation Examples from Mapillary Vistas.} In panoptic segmentation results, different instances are color-coded with different colors with small variations from the base color of their semantic class. In mismatched segments, segments belongs to true positives are marked as white, while false positive and false negative segments are marked as black.}
\label{fig:vistas-results}
\end{figure*}

In this appendix, we provide \ac{PQ} performance in detail on both Cityscapes (Table~\ref{tab:ciyscapes}) and Mapillary Vistas (Table~\ref{tab:vistas}) as well as more qualitative results (Figure~\ref{fig:cityscapes-results} and Figure~\ref{fig:vistas-results}).
These results were obtained using the test-time augmentation described in the main paper (multi-scale and flip).

We note that, in Cityscapes, some of the stuff classes that achieve the lowest \ac{PQ} scores (e.g. wall and fence) actually perform respectably on IoU. This is because \ac{PQ} treats \textit{all pixels} from each stuff class as a \textit{single} segment. 

This effectively penalizes stuff segmentation more harshly than things segmentation (where there are potentially multiple segments per class). We believe it's important to improve on the \ac{PQ} metric to strike a better balance between how thing and stuff classes are evaluated.

We can tell from Table~\ref{tab:vistas} that our main challenge on Mapillary Vistas is poor performance on rare classes, even though we used a weighted, bootstrapped loss~\cite{reed2014BCE} for training. Similar issues have been observed by existing approaches to semantic segmentation on this dataset~\cite{liulsun}. In future work we hope to explore other techniques for balanced batch sampling and rare class bootstrapping to improve TASCNet performance.


The examples in Figure~\ref{fig:cityscapes-results} and Figure~\ref{fig:vistas-results} further depict how our proposed method is able to achieve high quality panoptic segmentation using a unified architecture with a small backbone. 

Some examples also further illustrate our discussion of stuff bias in \ac{PQ}. In the third sample in Figure~\ref{fig:vistas-results}, although we correctly classify a substantial fraction of lane marking pixels, the lane marking \ac{PQ} for this image is $0$ as the fraction is under $50\%$.

\begin{table*}[b!]
\centering
\resizebox{0.38\textwidth}{!}{%
\begin{tabular}{@{}lcccc@{}}
\toprule
\textbf{Class} & \textbf{PQ} & \textbf{SQ} & \textbf{RQ}& \textbf{IoU} \\ \midrule
mean & 60.4 & 80.3 & 73.8&78.0 \\
\hline
road & 97.7 & 98.0 & 99.7&98.1 \\
\hline
sidewalk & 75.3 & 83.8 & 89.9 &85.1\\
\hline
building & 87.8 & 89.4 & 98.2 &92.5 \\
\hline
wall & 25.5 & 71.2 & 35.8& 55.8\\
\hline
fence & 27.0 & 71.5 & 37.7& 57.8\\
\hline
pole & 48.3 & 64.6 & 74.8 &64.3\\
\hline
traffic-light & 44.7 & 68.6 & 65.2 &71.2\\
\hline
traffic-sign & 66.1 & 75.9 & 87.0& 78.7\\
\hline
vegetation & 88.6 & 90.3 & 98.2 & 92.5\\ \bottomrule
\end{tabular}
}
\hspace{30pt}
\resizebox{0.38\textwidth}{!}{%
\begin{tabular}{@{}lcccc@{}}
\toprule
\textbf{Class} & \textbf{PQ} & \textbf{SQ} & \textbf{RQ}& \textbf{IoU} \\ \midrule
terrain & 28.9 & 73.4 & 39.3 &64.4\\
\hline
sky & 86.2 & 92.3 & 93.4 &95.1\\
\hline
person & 51.6 & 76.9 & 67.0&83.0 \\
\hline
rider & 53.5 & 71.7 & 74.6& 68.0\\
\hline
car & 64.8 & 83.2 & 77.8&93.6 \\
\hline
truck & 53.3 & 86.3 & 61.7&65.5 \\
\hline
bus & 72.9 & 88.5 & 82.4&87.7 \\
\hline
train & 65.6 & 83.1 & 78.9 &83.5\\
\hline
motorcycle & 44.5 & 72.3 & 61.5&68.4 \\
\hline
bicycle & 42.2 & 71.6 & 58.9&76.0 \\ \bottomrule
\end{tabular}
}

\caption{Panoptic Quality (PQ) on Cityscapes.\label{tab:ciyscapes} \ac{IoU}s are also included. Note that PQ, particularly for stuff classes, is a very stringent metric.}
\end{table*}

\begin{table*}[ht!]
\centering
\resizebox{0.45\textwidth}{0.45\textwidth}{%
\begin{tabular}{@{}lccc@{}}
\toprule
\textbf{Class} & \textbf{PQ} & \textbf{SQ} & \textbf{RQ} \\ \midrule
mean & 34.3 & 74 & 43.5 \\

animal--bird & 0.0 & 0.0 & 0.0 \\
animal--ground-animal & 39.1 & 80.5 & 48.5 \\
construction--barrier--curb & 42.9 & 70.5 & 60.9 \\
construction--barrier--fence & 32.0 & 72.2 & 44.3 \\
construction--barrier--guard-rail & 29.5 & 73.2 & 40.3 \\
construction--barrier--other-barrier & 24.3 & 77.4 & 31.3 \\
construction--barrier--wall & 19.4 & 72.7 & 26.7 \\
construction--flat--bike-lane & 13.0 & 69.9 & 18.7 \\
construction--flat--crosswalk-plain & 32.9 & 75.3 & 43.7 \\
construction--flat--curb-cut & 3.3 & 62.0 & 5.4 \\
construction--flat--parking & 5.9 & 64.2 & 9.2 \\
construction--flat--pedestrian-area & 19.1 & 85.2 & 22.4 \\
construction--flat--rail-track & 11.9 & 70.5 & 16.9 \\
construction--flat--road & 83.1 & 89.1 & 93.3 \\
construction--flat--service-lane & 28.6 & 80.2 & 35.7 \\
construction--flat--sidewalk & 52.4 & 77.6 & 67.6 \\
construction--structure--bridge & 33.1 & 76.6 & 43.2 \\
construction--structure--building & 69.2 & 83.0 & 83.5 \\
construction--structure--tunnel & 8.9 & 56.8 & 15.7 \\
human--person & 54.7 & 78.7 & 69.5 \\
human--rider--bicyclist & 43.3 & 73.6 & 58.8 \\
human--rider--motorcyclist & 38.2 & 70.3 & 54.3 \\
human--rider--other-rider & 7.8 & 52.8 & 14.8 \\
marking--crosswalk-zebra & 46.1 & 76.1 & 60.5 \\
marking--general & 41.1 & 68.6 & 59.9 \\
nature--mountain & 22.1 & 71.8 & 30.8 \\
nature--sand & 4.2 & 81.5 & 5.1 \\
nature--sky & 96.0 & 96.8 & 99.2 \\
nature--snow & 33.8 & 82.3 & 41.1 \\
nature--terrain & 37.8 & 76.9 & 49.2 \\
nature--vegetation & 81.3 & 86.0 & 94.5 \\
nature--water & 17.7 & 75.4 & 23.4 \\
\bottomrule
\end{tabular}
}
\hspace{10pt}
\resizebox{0.45\textwidth}{0.45\textwidth}{%
\begin{tabular}{@{}lccc@{}}
\toprule
\textbf{Class} & \textbf{PQ} & \textbf{SQ} & \textbf{RQ} \\ \midrule
object--banner & 30.6 & 82.3 & 37.1 \\
object--bench & 26.8 & 73.8 & 36.4 \\
object--bike-rack & 6.7 & 71.2 & 9.4 \\
object--billboard & 40.1 & 82.1 & 48.8 \\
object--catch-basin & 35.7 & 75.4 & 47.4 \\
object--cctv-camera & 21.0 & 69.5 & 30.2 \\
object--fire-hydrant & 56.8 & 80.8 & 70.2 \\
object--junction-box & 41.1 & 85.4 & 48.1 \\
object--mailbox & 23.0 & 79.0 & 29.2 \\
object--manhole & 47.7 & 81.2 & 58.8 \\
object--phone-booth & 15.1 & 84.7 & 17.9 \\
object--pothole & 0.6 & 63.0 & 1.0 \\
object--street-light & 46.0 & 74.0 & 62.1 \\
object--support--pole & 33.8 & 70.3 & 48.0 \\
object--support--traffic-sign-frame & 18.8 & 64.4 & 29.2 \\
object--support--utility-pole & 36.9 & 68.4 & 53.9 \\
object--traffic-light & 59.0 & 79.4 & 74.4 \\
object--traffic-sign--back & 39.0 & 74.9 & 52.2 \\
object--traffic-sign--front & 58.8 & 83.8 & 70.2 \\
object--trash-can & 48.5 & 83.7 & 57.9 \\
object--vehicle--bicycle & 37.7 & 72.1 & 52.2 \\
object--vehicle--boat & 17.1 & 67.8 & 25.3 \\
object--vehicle--bus & 54.6 & 87.6 & 62.3 \\
object--vehicle--car & 68.7 & 85.8 & 80.1 \\
object--vehicle--caravan & 0.0 & 0.0 & 0.0 \\
object--vehicle--motorcycle & 45.0 & 73.1 & 61.6 \\
object--vehicle--on-rails & 6.9 & 79.4 & 8.7 \\
object--vehicle--other-vehicle & 22.2 & 74.6 & 29.7 \\
object--vehicle--trailer & 13.6 & 79.3 & 17.1 \\
object--vehicle--truck & 51.6 & 86.3 & 59.8 \\
object--vehicle--wheeled-slow & 27.8 & 70.3 & 39.5 \\
void--car-mount & 51.7 & 85.2 & 60.7 \\
void--ego-vehicle & 71.5 & 90.4 & 79.2 \\ \bottomrule
\end{tabular}
}

\caption{Panoptic Quality (PQ) on Mapillary Vistas.\label{tab:vistas}}
\end{table*}

\end{document}